%% file: main.tex
\documentclass[sigconf]{acmart}

\renewcommand\footnotetextcopyrightpermission[1]{} % removes footnote with conference information in first column
\usepackage{flushend}
\input{preamble}

\begin{document}

%TODO: Replace the title with your project title.
\title{Enabling All In-Edge Deep Learning: A Literature Review}

\author{Praveen Joshi\textsuperscript{\rm 1}\footnotemark, Mohammed Hasanuzzaman\textsuperscript{\rm 1}, Chandra Thapa\textsuperscript{\rm 2}, Haithem Afli\textsuperscript{\rm 1},  and Ted Scully\textsuperscript{\rm 1}}
\affiliation{\textsuperscript{\rm 1}Munster Technological University,  Rossa Ave, Bishopstown, Cork, T12 P928, Ireland}
\affiliation{\textsuperscript{\rm 2}CSIRO Data61, Sydney, Australia}

% \corresp{Corresponding author: Praveen Joshi (e-mail: praveen.joshi@mycit.ie).}
\thanks{*Corresponding author: Praveen Joshi (e-mail: praveen.joshi@mycit.ie).}

\begin{abstract}
In recent years, deep learning (DL) models have demonstrated remarkable achievements on non-trivial tasks such as speech recognition, image processing, and natural language understanding. One of the significant contributors to its success is the proliferation of end devices that acted as a catalyst to provide data for data-hungry DL models. However, computing DL training and inference is the main challenge. Usually, central cloud servers are used for the computation, but it opens up other significant challenges, such as high latency, increased communication costs, and privacy concerns. To mitigate these drawbacks, considerable efforts have been made to push the processing of DL models to edge servers (a mesh of computing devices near end devices). Moreover, the confluence point of DL and edge has given rise to edge intelligence (EI). International Electrotechnical Commission (IEC) defines EI as the concept where the data is acquired, stored, and processed utilizing edge computing with DL and advanced networking capabilities.
Broadly, EI has six levels of categories based on where the training and inference of DL take place, \emph{e.g.}, cloud server, edge server and end devices. This survey paper focuses primarily on the fifth level of EI, called \emph{all in-edge} level, where DL training and inference (deployment) are performed solely by edge servers. All in-edge is suitable when the end devices have low computing resources, \emph{e.g.}, Internet-of-Things, and other requirements such as latency and communication cost are important such as in mission-critical applications (\emph{e.g.}, health care). Besides, 5G/6G networks are envisioned to use all in-edge.     
%
% As EI utilizes the computational power available at the cloud, edge, and end devices for DL training and deployment, it is broadly categorized into six levels. 
% However, this survey paper focuses primarily on the fifth level, the all in-edge level. The all in-edge level takes responsibility for training and deployment of the DL model solely on ESs, which forms the foundation for this survey paper. 
%
Firstly, this paper presents all in-edge computing architectures, including centralized, decentralized, and distributed. Secondly, this paper presents enabling technologies, such as model parallelism, data parallelism, and split learning, which facilitates DL training and deployment at edge servers. Thirdly, model adaptation techniques based on model compression and conditional computation are described because the standard cloud-based DL deployment cannot be directly applied to all in-edge due to its limited computational resources. Fourthly, this paper discusses eleven key performance metrics to evaluate the performance of DL at all in-edge efficiently. Finally, several open research challenges in the area of all in-edge are presented.
% Finally, open research challenges and future work directions to enable all in-edge are discussed for the benefit of the research community.
\end{abstract}
\maketitle

\section{Introduction}
\label{sec:introduction}

\textbf{The} global community is increasingly becoming a data-driven environment in which end devices are generating vast quantities of data outside of the traditional data centers. International Telecommunication Union anticipates that global internet traffic per month will reach 607 Exabytes (EB) in 2025 and 5016 EB in 2030 \cite{ITUR}. This enormous amount of data has a positive impact on artificial intelligence (AI) applications. In particular, deep learning (DL) relies on the availability of large quantities of data for its development, including training and inference~\cite{aggarwal2022has,lv2022look}.

DL has shown promising progress in natural language processing, computer vision, and big data analysis in recent years. For example, DL models, such as BERT, Megatron-LM, GPT-3, and Gropher, are reaching a human-level understanding of the textual data in natural language processing tasks~\cite{dale2021gpt}. Moreover, DL models have exceeded human performance on various tasks, including object classification tasks~\cite{barbu2019objectnet,kwabena2022capsule} and real-time strategy games~\cite{wu2019hierarchical}.

DL training and deployment in the majority of scenarios use a centralized cloud-based structure.
% A centralized cloud-based structure has been typically used to facilitate DL training as well as deployment in the majority of scenarios. 
However, the need to collect, process, and transfer vast data to the central cloud often becomes a bottleneck in many mission-critical use cases \cite{ray2019EDGE,sulieman2022EDGE}. In this regard, edge computing provides a high-performance bridge from local systems to private and public clouds.
%
%Edge computing provides a high-performance bridge from local systems to private and public clouds to accommodate the drawback of the centralized cloud-based design. 
%
The edge of the network, which often has modest hardware and memory resources (depending on the network infrastructure provider), can offer vital infrastructure to facilitate DL at the edge. Traditionally to avoid the bottleneck in many mission-critical use cases, edge computing performs tasks such as collection, filtering, and lightweight computation of raw data before transferring data to the cloud \cite{alsalemi2022innovative}. However, with the proliferation of edge servers and progress in DL-based architectures and algorithms, there is a possibility to perform DL model training and deployment efficiently at the network's edge.

The convergence of DL and edge computing has given rise to a new paradigm of intelligence called edge intelligence (EI)~\cite{wang2019EDGE,han2015learning}. EI aims to facilitate DL deployment closer to the data-generating source. As EI exploits the full potential of resources available at end devices, edge servers, and cloud servers for DL training and inference, based on resource utilization, it is categorized into six levels~\cite{zhou2019EDGE}. These six levels are defined based on where the DL-model training is taking place and where it is getting deployed in the network hierarchy. For simplicity, we assume a network hierarchy formed of cloud servers, edge servers, and end devices. DL training and deployment at cloud servers face significant challenges, including issues such as high latency, data privacy, network congestion, and security threats such as Denial-of-Service attacks~\cite{sadeeq2021iot}. On the other hand, despite being available in huge quantities, end devices suffer from constrained computation power, which is particularly relevant in the context of training and deployment of large DL models. In this regard, edge servers are a viable alternative.
%to address the challenges originating cloud servers and end devices. 
%
Moreover, due to their closer proximity to end devices, edge servers enable reducing network congestion in comparison to the centralized cloud architecture. 
% which mitigates the congestion faced in centralized cloud architectures. 
Furthermore, this proximity minimizes latency, providing for quicker inference when compared with DL models deployed at the cloud server. Even though edge servers have less computing power than the cloud, they do have significantly more computational power than end devices. Thus, edge servers can train and deploy DL models that require larger computing resources than that available at the end devices. 
%which is hard to achieve with the same number of end devices. 
%

The exclusive use of edge servers for both DL training and deployment is called \emph{all in-edge}.
% , and it is the focus of this survey paper. % 
% As the all in-edge level is composed of only ESs, it's basically can be seen as an ecosystem of itself with qualities such as low latency and faster inference than cloud level and modestly high computing infrastructure than end devices (which are not limited to small computers but also include IoTs with minimal computing power). 
%
Innovations and research on the emerging area of all in-edge DL processing are in their infancy. Unlike prior surveys \cite{deng2020EDGE,wang2020convergence,chen2019deep,park2019wireless,zhou2019EDGE,murshed2021machine} summarized in Table~\ref{tab:summary}, to the best of our knowledge, none of the existing surveys presents a detailed view from the all in-edge level perspectives on its enablers, key metrics of performances and challenges when DL is processed at all in-edge level. 
Specifically, this survey answers the following:
\noindent
\begin{enumerate}
\item Which architecture (centralized, decentralized, and distributed) should be used if the configuration of edge servers is known at the all in-edge level?
\item What are the state-of-the-art enabling technologies that facilitate DL training and inference from the all in-edge level?
\item What are the critical performance metrics required in addition to the standard metrics (\emph{e.g.}, accuracy and precision) to evaluate the performance of the DL model's applications at the all in-edge level?
\end{enumerate}

\begin{table*}[]
\centering
\caption{A summary of related surveys.\\ \ding{55}: Not included; \ding{108}: Not considered from all in-edge paradigm; \ding{52}: Included}
\label{tab:summary}
\resizebox{\textwidth}{!}{%
\begin{tabular}{|l|l|c|c|c|c|}
\hline  \hline
\multicolumn{1}{|c|}{\textbf{Survey Paper}}                                                                                                   &  \multicolumn{1}{c|}{\textbf{Takeaway}}                                                                                           

& \begin{tabular}[c]{@{}c@{}}\textbf{Discussion on} \\ \textbf{computing paradigm}\end{tabular}

& \begin{tabular}[c]{@{}c@{}}\textbf{Focused on all in-edge} \\ \textbf{Enablers}\end{tabular} 

& \begin{tabular}[c]{@{}c@{}}\textbf{Focused on all in-edge} \\ \textbf{Model Adaption}\end{tabular} 

& \begin{tabular}[c]{@{}c@{}}\textbf{Focused on all in-edge} \\ \textbf{Evaluation Metrics} \end{tabular} 
 \\ \hline  \hline
\begin{tabular}[c]{@{}l@{}}Edge intelligence: the confluence\\ of edge computing and \\ artificial intelligence~\cite{deng2020EDGE}\end{tabular}         & \begin{tabular}[c]{@{}l@{}}Survey provided insights into edge \\ intelligence. It partitioned edge \\ Intelligence into AI for edge and AI\\  on edge along with research roadmap.\end{tabular}                                                    & \ding{55}                                                               & \ding{108}                                                                             & \ding{108}                                                                                & \ding{55}                                                                                                                                                              \\ \hline
\begin{tabular}[c]{@{}l@{}}Convergence of edge computing\\ and deep learning: \\ A comprehensive survey~\cite{wang2020convergence}\end{tabular}                 & \begin{tabular}[c]{@{}l@{}}Survey looked upon EI from \\ machine learning perspective for wireless \\ communication. Also, provides insights \\ into the edge hardware for DL.\end{tabular}                                         & \ding{108}                                                            & \ding{108}                                                                           & \ding{108}                                                                                 & \ding{55}                                                                                                                                                             \\ \hline
\begin{tabular}[c]{@{}l@{}}Deep Learning with edge \\ computing: A review~\cite{chen2019deep}\end{tabular}                                               & \begin{tabular}[c]{@{}l@{}}Authors of the survey paper looked upon\\ scenarios of EI, techniques to speed up\\  training and inference on ED.\end{tabular}                                                                              & \ding{55}                                                             & \ding{108}                                                                           & \ding{108}                                                                                & \ding{55}                                                                                                                                                             \\ \hline
\begin{tabular}[c]{@{}l@{}}Wireless network intelligence\\  at the edge~\cite{park2019wireless}\end{tabular}                                                 & \begin{tabular}[c]{@{}l@{}}Survey provided insights into theoretical \\ and technical enabler of edge ML for\\ training and inference process.\end{tabular}                                                                                        & \ding{55}                                                            & \ding{108}                                                                           & \ding{108}                                                                                & \ding{55}                                                                                                                                                             \\ \hline
\begin{tabular}[c]{@{}l@{}}Machine Learning at the \\ Network edge: A Survey~\cite{murshed2021machine}\end{tabular}                                            & \begin{tabular}[c]{@{}l@{}}Survey looks upon the deployment of ML\\  system at edge of computer along with\\  the tools, frameworks and hardware.\end{tabular}                                                                                     & \ding{55}                                                              & \ding{108}                                                                           & \ding{108}                                                                                & \ding{55}                                                                                                                                                            \\ \hline
\begin{tabular}[c]{@{}l@{}}Edge Intelligence: Paving \\ the Last Mile of Artificial \\ Intelligence With edge\\ Computing~\cite{zhou2019EDGE}\end{tabular} & \begin{tabular}[c]{@{}l@{}}Authors provides six level rating for EI. \\ Survey also provides into the architecture, \\ framework and technologies require for\\  DL deployment  over edge.\end{tabular}                                            & \ding{52}                                                              & \ding{108}                                                                           & \ding{108}                                                                                & \ding{108}                                                                                                                                                           \\ \hline
\begin{tabular}[c]{@{}l@{}}Enabling All In-Edge Deep Learning:\\ A
Literature Review \\ (Ours)\end{tabular}                                  & \begin{tabular}[c]{@{}l@{}}Survey looks upon the key architecture, \\ enabling technologies, model adaption \\ techniques along with performance metric \\ for training and inference from DL for \\ all in-edge level.\end{tabular} & \ding{52}                                                               & \ding{52}                                                                            & \ding{52}                                                                                 & \ding{52}                                                                                                                                                              \\ \hline  \hline
\end{tabular}
}
\end{table*}

\begin{figure*}[h]
\centering 
\setlength
\fboxsep{0pt} 
\setlength\fboxrule{0.25pt} 
\fbox{\includegraphics[width=6.2in]{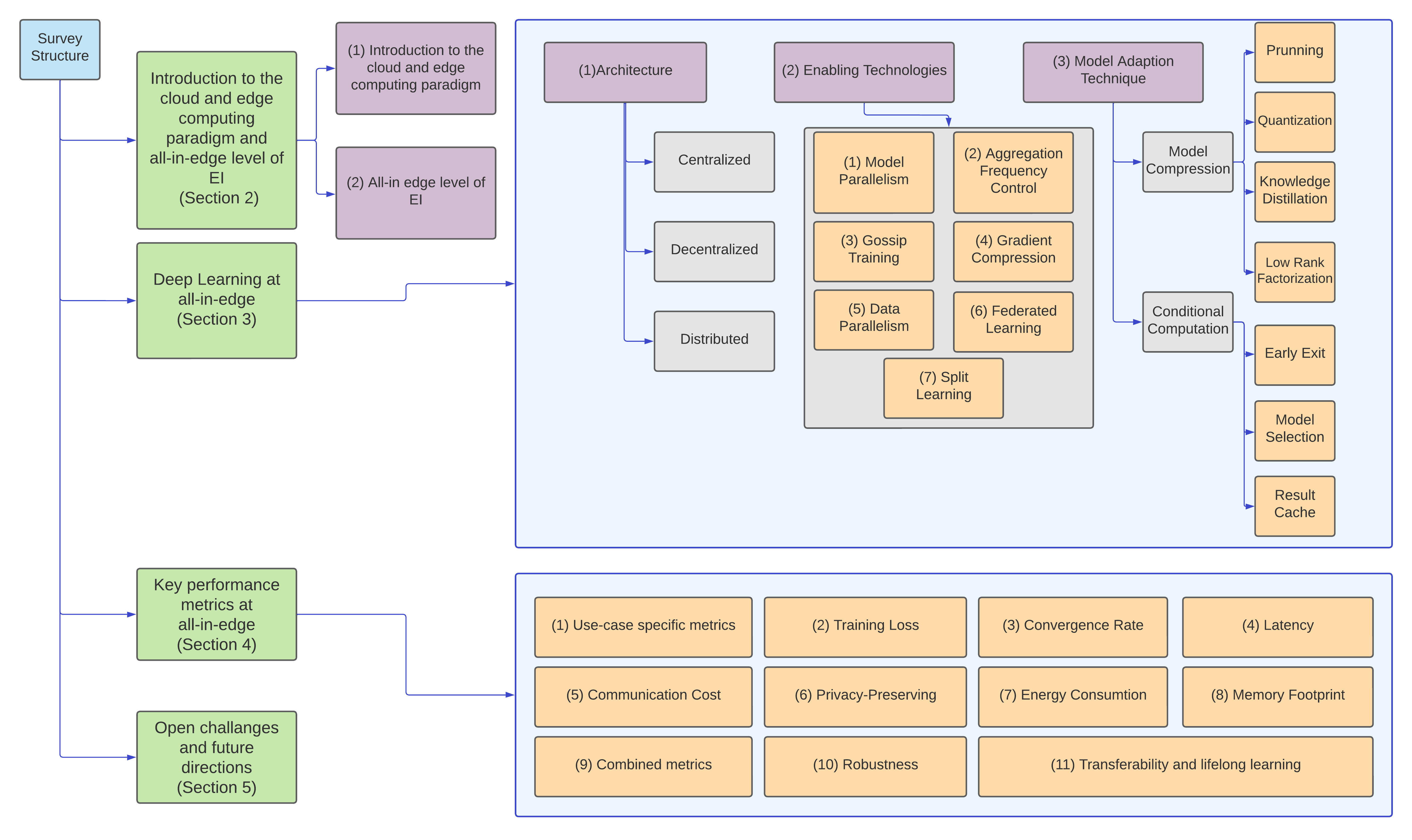}} \caption{ An overview of the structure of the survey paper.} \label{overview} \end{figure*}

% An overview of the survey paper organisation is provided in Figure \ref{overview}. The survey paper is organised as follows:
% \begin{enumerate}
% \iffalse
% \item Section I provides an insight into the survey paper, gaps in the existing survey and contributions by current survey paper.
% \fi
%     \item Section II provides the primer on computing paradigm and EI. This section also defines the all in-edge paradigm of EI.
%     \item Section III presents the architecture, enabling technologies for training and inference from the DL models at the all in-edge paradigm, and also examines model adaption technique for effective deployment of DL models at edge.
%     \item Section IV reviews the key performance metrics used for evaluating the research in all in-edge DL.
%     \item Section V discusses the open challenges and future direction of research in all in-edge DL.
%     \item Section VI presents a summary and identifies the primary conclusions and findings of the paper.
% \end{enumerate}

This paper is organized in the following way. It first introduces the computing paradigm and the all in-edge level of EI in Section~{II}. Then, in Section~{III}, it discusses the architecture, enabling technologies for training and inference of DL models at the all in-edge paradigm. Besides, this paper examines the model adaption techniques for effectively deploying DL models at the edge. Next, it reviews the key performance metrics used for evaluating all in-edge DL processing in Section~{IV}. Section~{V} discusses the open challenges and future direction of research for DL at all in-edge. Finally, Section~{VI} presents a summary and identifies the primary conclusions and findings of the paper. Overall, Figure~\ref{overview} depicts the organization of this paper in the block diagram, and Table~\ref{tab:Abb} provides the list of important acronyms.

\begin{table}[h]
\begin{center}
\caption{List of important abbreviations}
\label{tab:Abb}
\setlength{\tabcolsep}{2pt}
\begin{tabular}{|p{85pt}|p{135pt}|}
\hline\hline
\textbf{Abbreviation}& 
\textbf{Definition}\\
\hline\hline
AI& 
Artificial Intelligence\\
\hline
ANN&
Artificial Neural Network\\
\hline
CC&	Cloud Computing\\
\hline
CIPAA&
Construction Industry Payment and Adjudication Act\\
\hline
DC&
Data Center\\
\hline
DL&
Deep Learning\\
\hline
DNN&
Deep Neural Network\\
\hline
EC&
Edge Computing\\
\hline
EDs&
Edge Devices\\
\hline
EDGE&
Enhanced Data GSM Environment\\
\hline
EI&
Edge Intelligence\\
\hline
ES&
Edge Server\\
\hline
GDPR&
General Data Protection Regulation\\
\hline
IoT&
Internet of Things\\
\hline
MEC&
Mobile (Multi-Access) Edge Computing\\ \hline
QoS  & Quality of Service \\
\hline\hline
\end{tabular}
    
\end{center}
\label{tab1}
\end{table}

\section{Preliminary}

%his section introduces the cloud and edge computing paradigm and all in-edge levels of EI.
%
The centralized nature of the cloud data center has several drawbacks. One of the most considerable disadvantages is the distance between the data centers and end (user) devices, as it requires more wait time to process the data. On the other hand, edge computing offers an indisputable advantage by physically moving storage and processing resources closer to the source of data generation, thereby achieving lower latency. This section presents the distinction between the cloud and edge computing paradigms. Besides, it presents the all in-edge level of the EI paradigm, which comprises only edge servers.
\iffalse
\begin{figure*}[h]
\centering 
\setlength
\fboxsep{0pt} 
\setlength\fboxrule{0.25pt} 
\fbox{\includegraphics[width=6.2in]{image/N_EDGELayer.png}} \caption{Illustration of Cloud-Edge computing-based IoT.} \label{fig1} \end{figure*} 
\fi

\subsection{Introduction to the cloud and edge computing paradigm}
The computation of DL can be done by various devices, including cloud servers, edge servers (ESs) and edge devices(EDs). This determines the following computing paradigms. 
%
%A significant computational and storage capacity is required to process DL. As such, these requirements cause substantive impediments to the deployment of DL models on IoT and other resource constrained devices. 
%Below we evaluate the cloud and edge computing paradigms in the context of DL based on their storage, computational power, and proximity to the ED. 

\subsubsection{Cloud Computing}
Cloud computing is a paradigm for wide-reaching distributed computing that uses technologies such as grid computing, service orientation, and virtualization. It enables on-demand infrastructure access to a shared pool of configurable computing resources that can be acquired and released with minimum intervention from the server infrastructure provider. Cloud servers have significant storage capacity and computational power to facilitate the overwhelming data coming via the backhaul network from end-user \cite{salem2020ai,chen2016Cloud}. Thus, cloud servers can satisfy resource requirements for aggregation, pre-processing, and inference for any artificial intelligence-based applications. The cloud servers are inter-connected, providing global coverage with a backhaul network. The cloud computing paradigm involves the end devices that offload data directly to the cloud for further processing. The end devices mentioned here are the originators of the data. In the cloud, data can persist for days, months, and years, meaning long-term temporal data can be collated and processed. For example, cloud data centers facilitate forecasting models based on a large amount of historical time series data \cite{puliafito2019fog}. 
\iffalse
With the proliferation of IoT in the prevailing world, data generated have been exponentially rising. To provide almost real-time prediction, the cloud holds the powerful computing resources necessary for processing data.
\fi
Cloud computing is still the appropriate vehicle for modeling and analytical processing if latency requirements and bandwidth consumption are not an issue, provided measures for preserving privacy and security are in place \cite{dantas2020application}.

\subsubsection{Edge Computing} 
With the surge in the proliferation of IoT devices, traditional centralized cloud computing struggles to provide an acceptable Quality of Service (QoS) level to the end customers \cite{ali2020deep}. To meet the QoS of IoT applications, there is a need for cloud computing services closer to data sources (\emph{e.g.}, IoT devices, EDs, etc.). As defined by  International Electrotechnical Commission (IEC), the extension of computing services from cloud computing to the network edge is called edge computing (EC)~\cite{barate20195g, kupriyanovsky2019eu}. Edge computing helps application developers cater to user-centric services closer to clients. In contrast to cloud computing, latency incurred from edge computing is significantly less, as a majority of data does not have to travel via a backhaul network to the cloud \cite{saleem2020latency}. Less consumption of backhaul networks also means the requirement of bandwidth consumption is considerably less, as shown in Figure \ref{fig2}. 

\subsection{All in-edge level of edge intelligence}
\label{EIAllInEdge}
Significant progress has been made in the DL domain in the last decade. Technical advancements in high-performance processors \cite{li2020survey} coupled with improvements in DL algorithms, and the availability and maturity of big data processing \cite{sanchez2016workshop} have contributed to the increase in DL performance. However, DL processing (training and inference) still occurs mainly in the cloud, as DL models require significant computational resources. As mentioned earlier, this can adversely impact the DL’s QoS due to high latency. At the same time, there has been substantial research focused on facilitating DL processing at the edge. While edge computing provides relatively modest computing resources and storage capacity, the training and deployment of DL applications on such devices would greatly help in achieving acceptable QoS for real-time DL applications. For example, real-time applications that would benefit from the merger between edge computing and DL include automated driving \cite{milz2018visual}, and real-time surveillance \cite{barthelemy2019EDGE}, all of which intrinsically require fast processing and rapid response time \cite{liu2018distributed,hassan2018role}. The concept of edge intelligence is a new paradigm that utilizes end devices, edge nodes, and cloud data centers to optimize the processing of DL models (for both training and inference) \cite{zhou2019EDGE}.\par
\begin{figure*}[h]
\centering 
\setlength
\fboxsep{0pt} 
\setlength\fboxrule{0.25pt} 
\fbox{\includegraphics[width=6.2in]{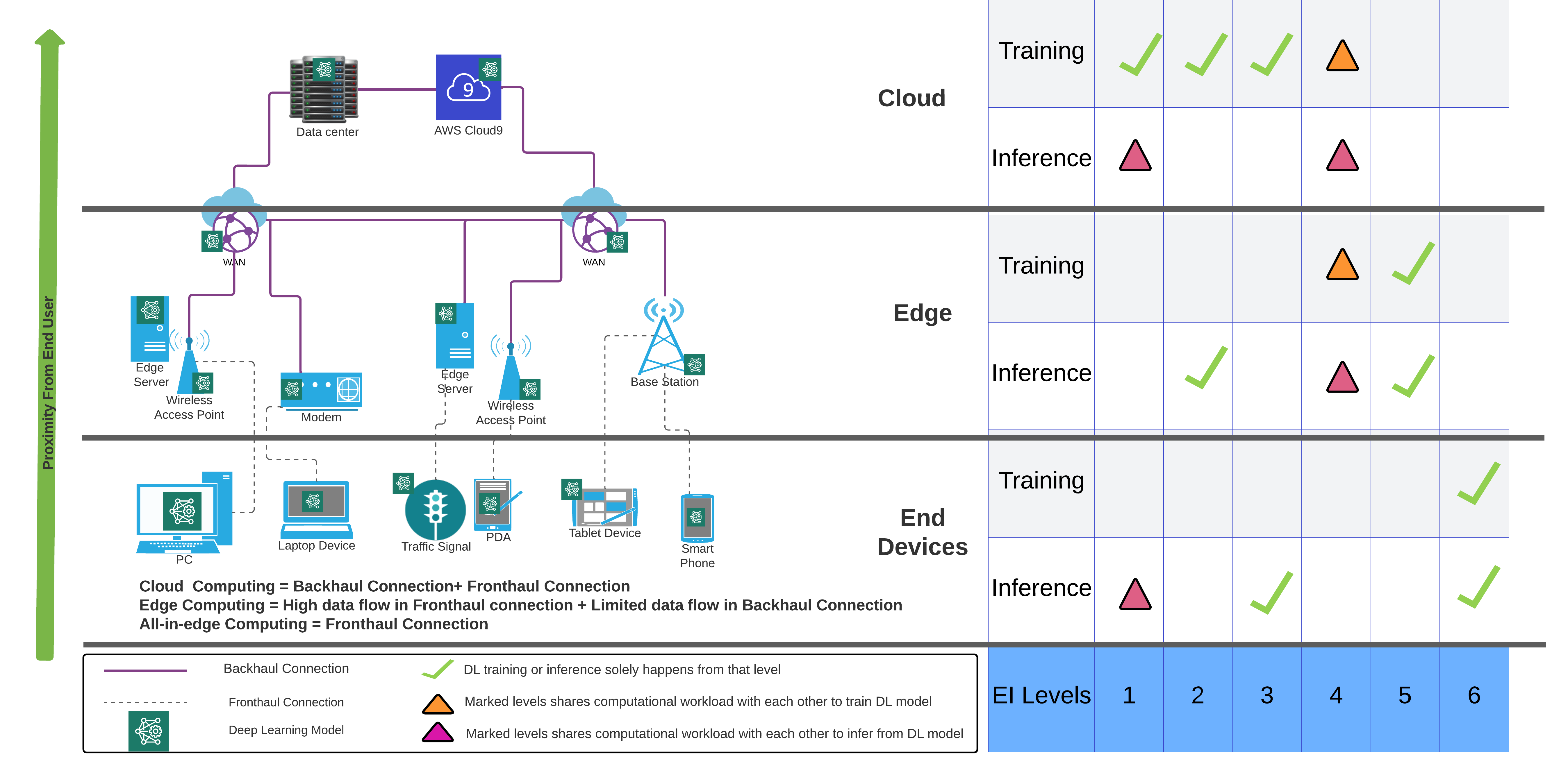}} 
\caption{A layered network architecture with cloud, edge and end devices (the left part of the figure), and the ratings of edge intelligence (EI) into six levels (the right part of the figure).} 
\label{fig2} 
\end{figure*}

  
As depicted in Figure \ref{fig2}, edge intelligence is divided into six distinct levels based on computational resources offered by the cloud, edge, and end devices for the DL training and inference phase. The fifth level of edge intelligence, depicted in Figure~\ref{fig2}, corresponds to all in-edge processing. As defined in~\cite{zhou2019EDGE}, all in-edge (fifth level) refers to the edge intelligence paradigm where both training and inference of the Deep Neural Network (DNN) take place in the ES (also known as in-edge manner). This level is critical to satisfying the latency requirements of real-time artificial intelligent applications. In addition, it is helpful in scenarios with intermittent or limited connectivity to the backhaul network \cite{oderhohwo2020EDGE}.
This level helps in reducing the amount of data that needs to be transferred from end devices to the cloud whenever the DL model is being trained. Also, inference provided by the all in-edge level is faster than any other level of EI where inference takes place in the cloud data center~\cite{charyyev2020latency}. 

The modest computational resources available with ESs when processing DL models at an all in-edge level facilitate training and inference of relatively large models \cite{huang2021enabling}. Based on the DL model's size, either a single ES can train a DL model or a group of ESs collaborate to train a DL model. Technologies for training DL at level five are described in detail in Section~\ref{EnablingTechnologies}. Similarly, inference from all in-edge can be produced from either a single ES or multiple ESs working collaboratively.

\iffalse
Edge computing now offers greater computing capability due to significant advancements in the computing power of ES and the microprocessors used in end devices \cite{mittal2020survey}. Consequently, this improvement in performance means that end devices in future will become less reliant on cloud servers for the provision of DL inference and training purposes. The all in-edge level (EI level 5th) attempts to facilitate the end user with the necessary computation, storage and latency requirements lessening the dependence on a centralized cloud server. This research paper primarily focuses on the all in-edge level.
\fi
\section{Deep Learning at all in-edge}
\iffalse
Data generated by end-user devices facilitates the provision of AI-based analytics. More specifically, this data enables us to train DL models, which can be used for real-time inference. Each of the enabling technologies and model adaption techniques stated in this section can be applied to other levels of EI. As authors, specifically, look for an all in-edge level because of its nature:
\begin{enumerate}
    \item Low data transfer: As data is transferred from one ES to another, which is comparatively low than data transferred between the end device and cloud.
    \item Availability of computational power: ES have comparatively higher resources to provide more computational power than end devices.
\end{enumerate}
\fi
This section reviews the current state of the art for training and adapting DL models from the all in-edge level perspective. Furthermore, the section details the different architectures employed for DL training within the all in-edge level.

\subsection{Architecture}
\label{modelTrainingArch}
\begin{figure*}[h]
\centering 
\setlength
\fboxsep{0pt} 
\setlength\fboxrule{0.25pt} 
\fbox{\includegraphics[width=6.2in]{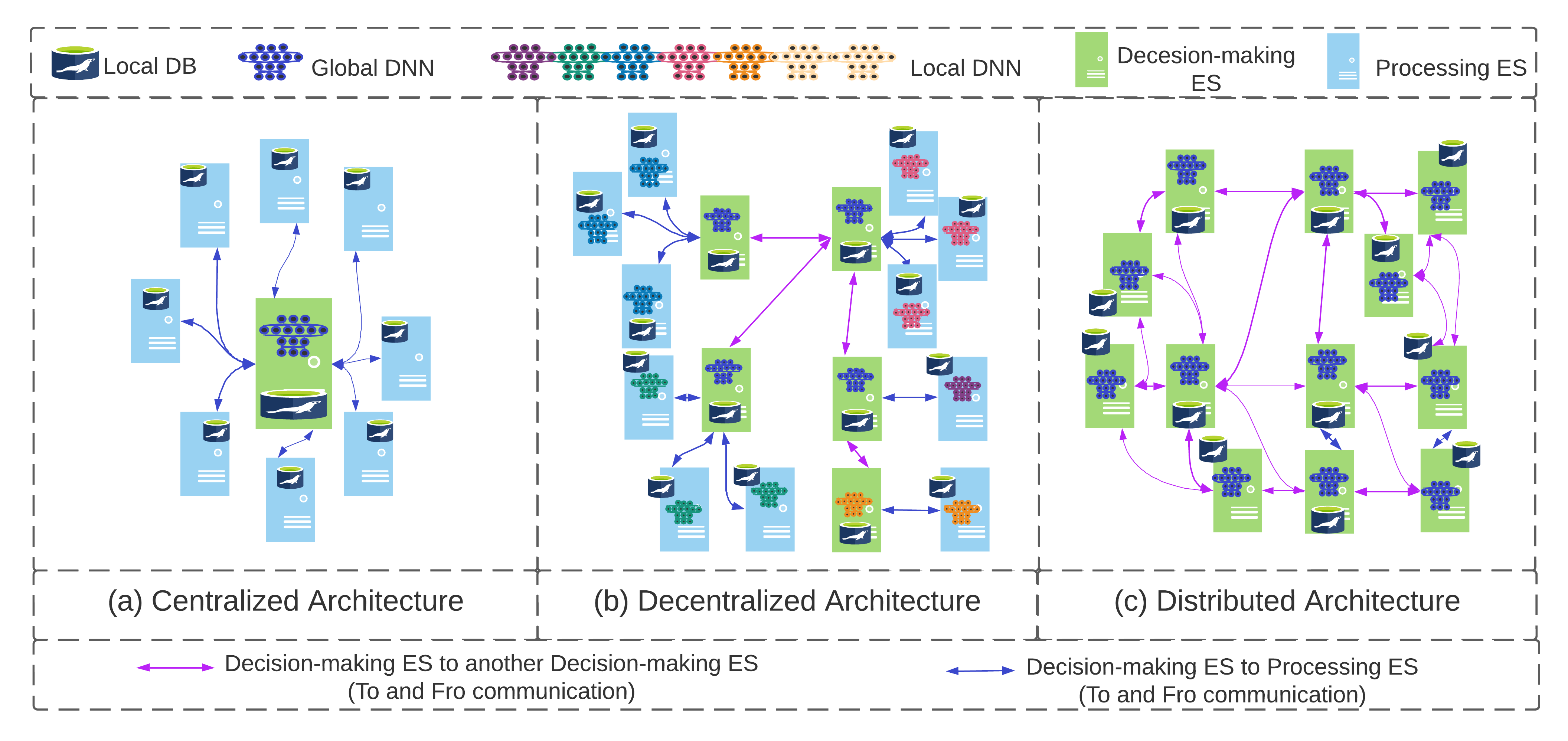}} \caption{Architecture for training Deep Learning model in-edge: (a) Centralized, (b) Decentralized, and (c) Distributed Architecture.} 
\label{fig3} 
\end{figure*}

The architecture used for DL training at the ES can broadly be divided into three main categories: centralized, distributed, and decentralized, as shown in Figure \ref{fig3}. The architecture is defined based on the role of two different types of ES. The first is the processing ES, which is tasked with training the DL model, and the second is the decision-making ES, which coordinates how the model is shared across the network.

%\begin{enumerate}[leftmargin=0.5cm]
\noindent
    \paragraph*{1) Centralized Architecture}
    In a centralized architecture (Figure \ref{fig3}(a)), the processing ES sends the data produced by the end devices (without training local DNN) to the decision-making ES. Decision-making ES then undertakes the DNN training task acting as processing ES \cite{rani2022cloud,kong2021real}. The centralized ES is assumed to have sufficient computing power (and typically, the computing power of the decision-making ES exceeds that of each of the processing ES). In this architecture, the decision-making ES is responsible for acting as both the processing and decision-making ES. Due to decision-making ES acting as processing ES at the same time makes it vulnerable to a single point of failure. 
 
 \noindent
    \paragraph*{2) Decentralized Architecture} 
    In a decentralized architecture, depicted in Figure \ref{fig3}(b), each processing ES is responsible for training its own local DNN. Once a local model is trained, the ESs send their local DNN model copy to a corresponding decision-making ES. This decision-making ES aggregates the DNN models and subsequently shares it with other decision-making ES to whom it is connected [40, 41]. Compared to centralized architecture, decentralized architecture addresses the single point of failure by dispersing models amongst multiple decision-making ESs. Thus, even if a single decision-making ES was to go offline, the system could continue operating.
    
    \begin{figure*}[]
    \centering 
    \setlength
    \fboxsep{0pt} 
    \setlength\fboxrule{0.25pt} 
    \fbox{\includegraphics[width=6.2in]{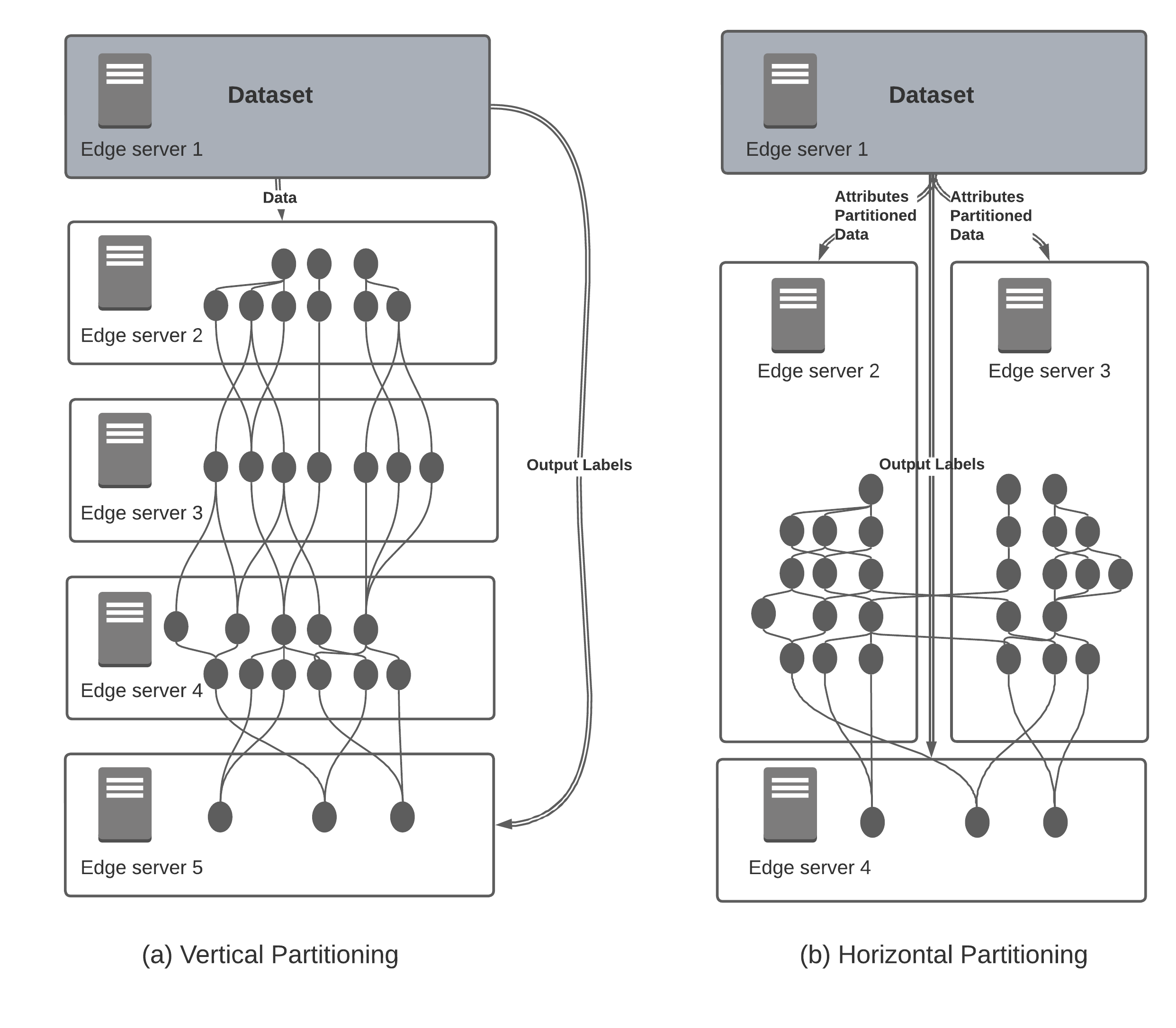}} \caption{Model Parallelism.} 
    \label{fig502} 
    \end{figure*} 
\noindent
    \paragraph*{3) Distributed Architecture} 
    A distributed architecture aims to provide a much more resilient architecture by making each ES (typically decision-making ES) capable of processing (training a local copy of DNN) and making decisions on how to share the data across the networks with other peers. In this architecture, each ES establishes a random peer-to-peer connection with another ES in the network for that specific iteration to share their local models. The receiving ES aggregates received model weights with their local copy of parameters. The training of DNN is stopped once the loss stabilizes in most ESs, and further updating the model parameters does not change the model's estimate for a given classification or regression problem~\cite{kong2021consensus}.
%\end{enumerate}

\subsection{All in-edge enabling technologies}
\label{EnablingTechnologies}
This section focuses on the technologies that enable the model training process undertaken by the ES. Model parallelism, aggregation frequency control, gossip training, gradient compression, data parallelism, federated learning, and split learning at the ES are emerging technologies, as seen by the substantial amount of research interest and citations, shown in Table~\ref{tab:ComparisonOfet}. 
 \paragraph*{1) Model Parallelism/DNN Splitting} 
 \label{lbl_model_partitioning}
 Model Parallelism (also referred to as model splitting or DNN Splitting) is a technique in which the DNN is split across multiple ESs to overcome the constrained computing resources. Model parallelism utilizes a decentralized architecture such that after DNN partitioning, a number of processing ESs train different layers of the DNN model, and a decision-making ES coordinate the training and ensure the correct flow of activations. The model partitioning ensures that the workload assigned to an individual processing ES does not exceed its computational capabilities. Model splitting can be categorized into either horizontally partitioned or vertically partitioned, as shown in Figure \ref{fig502}. In the vertical partitioning approach, one or more layers of the DNN are housed in different servers based on the computational requirement of the layer and the available resources of the processing ES. Whereas in horizontal partitioning, neurons from different layers are placed together based on the computational power of the processing ES. Horizontal partitioning is beneficial when input data is significantly big (number of attributes in a dataset) and single processing ES fails to perform a single-layer operation. 

In~\cite{kim2016strads}, the authors proposed a framework for scheduled model parallel machine learning called STRADS for vertical partitioned parallel machine learning. The DL application scheduler introduced in the STRADS framework helped control the update of the model parameters based on the model’s dependency structure and parameters of the DNN model. The authors also successfully demonstrated $10\times$ faster convergence of the model parallelism-based topic modeling implementation over the model without parallelism. In 2021, research \cite{narayanan2021efficient} on training the Megatron language model, authors utilized horizontally partitioned model parallelism to train a multi-billion parameter language model. In contrast to the single-GPU-per-model training, the authors in this research implemented model parallelism on the same PyTorch transformer implementations with few modifications. To train such a big system, 512 GPUs were consumed to train the transformer-based model. The same model was then able to achieve the SOTA accuracy on the ReAding Comprehension Dataset From Examinations (RACE~\cite{lai-etal-2017-race}) dataset with improved throughput by 10\% as compared to existing approaches. Model parallelism provides a way to combine the resources from multiple processing ES to enable all in-edge training of a single DL model.

\begin{figure}[h]
\centering 
\setlength
\fboxsep{0pt} 
\setlength\fboxrule{0.25pt} 
\fbox{\includegraphics[width=3.0in]{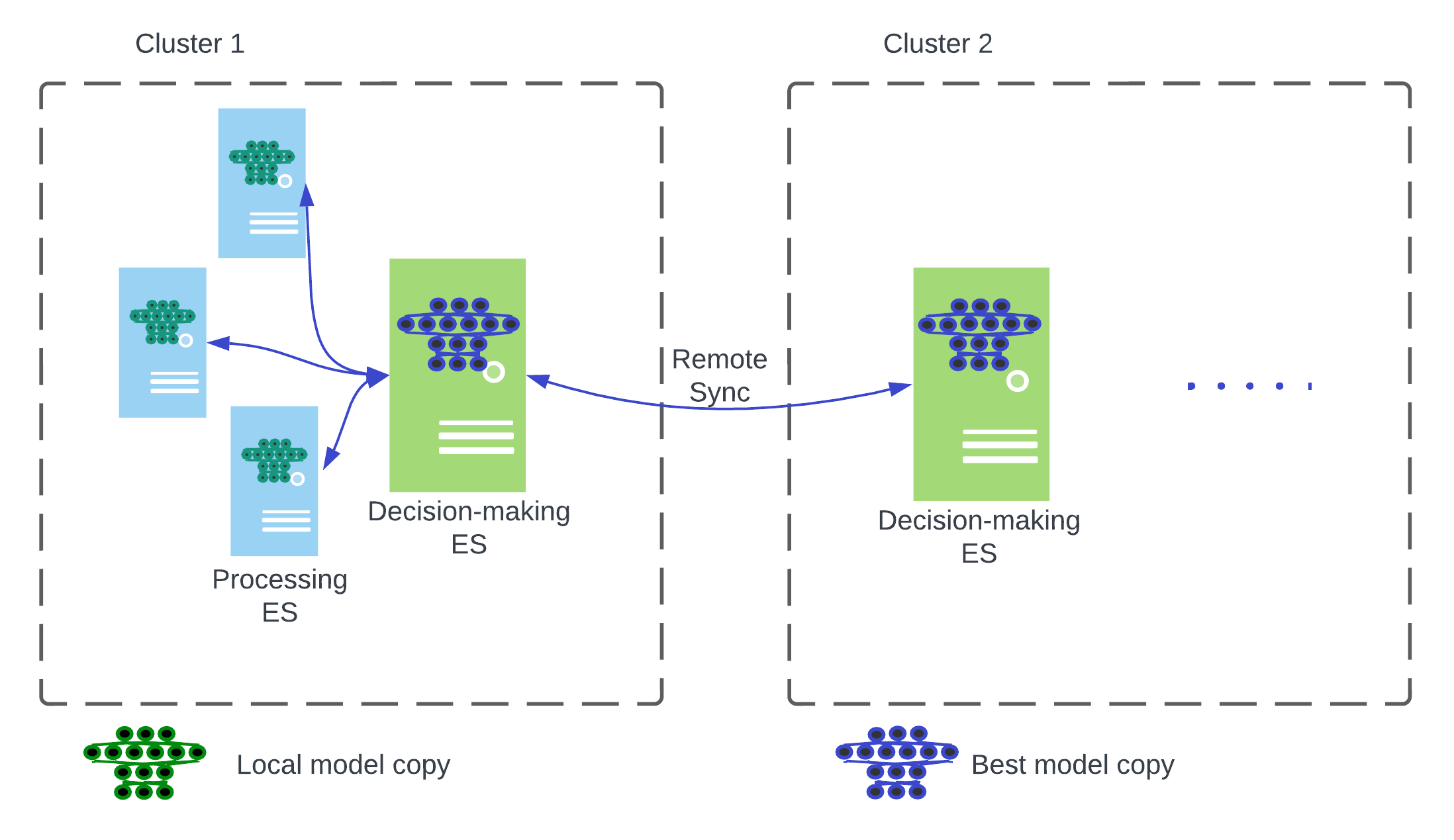}} \caption{Aggregated Frequency Control (AFC).} 
\label{fig6} 
\end{figure}

\paragraph*{2) Aggregated Frequency Control (AFC)}
AFC adopts a decentralized architecture for training DL models, in which a finite number of discrete clusters of ESs are formed, as shown in Figure \ref{fig6}. The task of each of the discrete clusters is to train an identical DNN model. Each cluster has one ES that acts as a decision-making ES. The task of the decision-making ES is to provide all processing ESs in the cluster with an identical copy of the DNN model. Once each processing ES receives its copy, they train that model using their local data and send back the updated DNN model weights to the decision-making ES for aggregation. The decision-making ES aggregates the weights from each of the individual processing ESs in the cluster. Once aggregation is done, the decision-making ES sends back the updated DNN model to all the processing ESs in the cluster. In addition, after each aggregation at the decision-making ES, a "significance function" is computed. This function will determine if the current aggregation has led to a significant improvement. If the improvement is deemed significant, then the current cluster's decision-making ES will inform the decision-making ES of each of the other clusters of the new model weights. Hence, each decision-making ES will have the best available model copy at any given point in time.

The significance function in AFC influences the frequency with which updated weights are sent from one decision-making ES to another. This, in turn, can reduce the communication overhead in the network. The Approximate Synchronous Parallel (ASP) model \cite{hsieh2017gaia} is one such model that targets the problem of geo-distributed DL training. This research successfully employed an intelligent communication system based on the AFC technique achieving minimization in WAN communication by the factor $1.8$-$53.5\times$ between the two data centers.
% By utilizing the AFC at all in-edge paradigm, one can benefit from the low communication between the parameter ES, which are far apart.\newline

\paragraph*{3) Gossip Training}  
\label{lbl:gossip}
Gossip Training provides a way to reduce the training time in a distributed architecture. Gossip training is based on the randomized selection of the ES to share the gradient weights for aggregation \cite{loizou2016new}. Each ES acts as a decision-making ES and processing ES to make the whole training system fault resilient. In this technique, ES will randomly select another node and subsequently send the gradient weight updates to the selected ES. Each ES will then compute the average received weights. Gossip training works in a synchronized and distributed manner. In \cite{blot2019distributed}, researchers demonstrated that GoSGD (Gossip Stochastic Gradient Descent) takes 43\% less time to converge to the same train loss score when compared to the EASGD (Elastic Averaging SGD \cite{zhang2015deep}) algorithm used in distributed architecture training.
In other research, PeerSGD \cite{vsajina2020decentralized} modified the GoSGD algorithm \cite{blot2019distributed} to work in the distributed trustless environment. The algorithm was modified at the stage when the random peer was selected to share the update. The peer who receives the update can decide whether to accept the received weights based on the loss difference (hyper-parameter defined in the research). PeerSGD was evaluated with various clients ranging from 1 to 100. In the experiment, PeerSGD demonstrated $2\times$ faster convergence when tested with 10 clients compared to 100 clients, but it still had comparable accuracy. The limitation of PeerSGD is its inability to achieve convergence in a scenario when data classes are segregated across multiple clients. A modified version of GoSGD was also applied to Wide Area Networks \cite{oguni2021communication}, and heterogeneous edge computing platforms \cite{han2020accelerating} and demonstrated results comparable to the original GoSGD algorithm. Gossip training facilitates all in-edge model training without any central authority, making the training process more resilient if any ES is not reachable during training.

\paragraph*{4) Gradient Compression} 
Gradient Compression is another approach to reducing communication while training the DL model, which can be applied to either a distributed or decentralized architecture to facilitate all in-edge training. Gradient compression minimizes the communication overhead incurred by addressing the issue of redundant gradients. Authors in the research \cite{han2018bandwidth} found that 99.9\% of the gradient exchange in distributed stochastic gradient descent is redundant. They proposed a technique called Deep Gradient Compression, which reduced the communication necessary for training ResNet-50 from 97 MB to 0.35 MB. In gradient compression, two approaches are used in practice: gradient quantization and gradient sparsification. \par
In gradient quantization  \cite{tang2018communication}, gradient weights are degraded from having a higher order of precision values to a lower precision order {\em i.e.}, representing weights using float 12 rather than float 64. In \cite{du2020high}, the author proposed high-dimensional stochastic gradient quantization for reducing the communication in the federated learning setting (federated learning setting is explained in~\ref{subsec_fedlearning}-6). In the proposed architecture, the authors utilized a uniform quantizer and low-dimensional Grassmannian to decompose the model parameters, followed by compression of the high-dimensional matrix of stochastic gradients into its norm and normalized block gradients. Normalized block gradients are then scaled with a hinge vector to yield the quantized normalized stochastic gradient (QNSD). This QNSD was then transmitted by the processing ES, who trained the model
to the decision-making ES, who then aggregates the various gradients and updates a global DL model. Through the framework of hierarchical gradient quantization, authors reduced the communication overhead theoretically and, at the same time, achieved a similar accuracy to the SOTA signSGD scheme\cite{bernstein2018signsgd}.  \par
\begin{figure}[h]
\centering 
\setlength
\fboxsep{0pt} 
\setlength\fboxrule{0.25pt} 
\fbox{\includegraphics[width=3.0in]{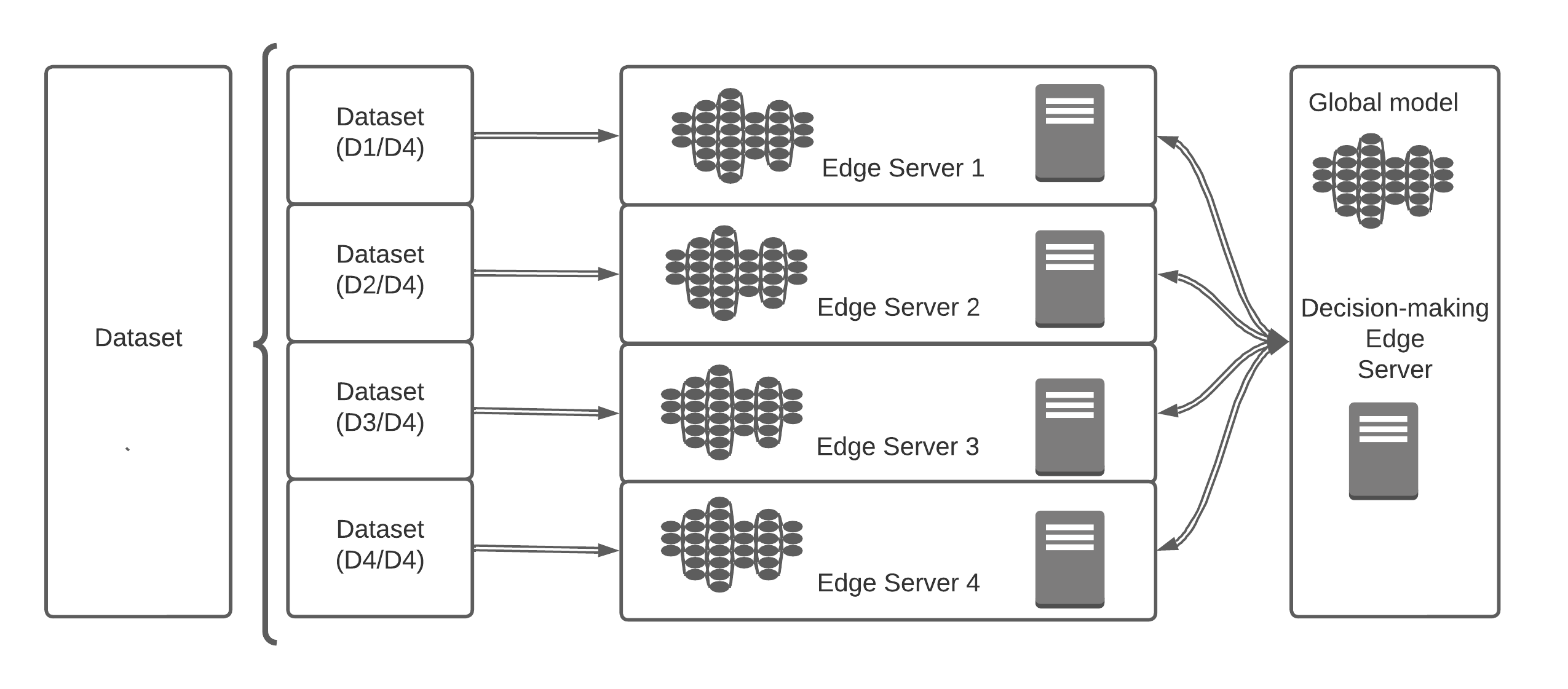}} \caption{Data Parallelism.} 
\label{fig501} 
\end{figure}

Another approach to gradient compression is gradient sparsification. This technique allows the gradient exchange only if the absolute gradient values are higher than a certain threshold \cite{strom2015scalable}. For example, the threshold in the research ranged from 2 to 15. So if the absolute values of the gradients elements exceed the threshold, they are allowed to be transmitted. The higher the value of the selected threshold, the lower the communication cost (as the threshold limits the transmission of gradient weights). This method reduced the required communication bandwidth by three orders of magnitude for data-parallel distributed SGD training of DNNs. Recent research \cite{tao2018esgd} found that selecting an appropriate threshold is challenging due to the variation in the value of the gradients. This research proposed an alternative approach called the edge Stochastic Gradient Descent (eSGD) method. In eSGD, determining if the gradient update should be sent over the network is based on the loss function.  The loss function is used to compute the loss against each coordinate of the gradient at time steps `$t-1$' and `$t$'. If the loss value at time step `$t$' is smaller than its value at time step `$t-1$', the current gradient `$gt$' will be transmitted to other ESs to build a global model. The standard SGD, when applied to MNIST with 128 batch size and trained for 200000 epochs, will achieve 99.7\% accuracy. In contrast, the eSGD method with the same setting attained an accuracy of 95.31\% and 91.22\% with a drop ratio (\% of gradients that will not be communicated by ES) of 25\% and 50\%, respectively. In \cite{shi2020communication}, the authors aim to identify an optimal trade-off between the communication that takes place within the layers of a DNN (housed in different ESs) and the computations required for the gradient sparsification. The authors developed an optimal merged gradient sparsification algorithm that required 31\% less time per iteration over the SOTA sparsified SGD. For the all in-edge paradigm, the size of the message being communicated by the servers utilizes a significant bandwidth. The gradient compression approach helps reduce the size of the message being communicated from one ES to another, thereby freeing up network bandwidth which can then be utilized by other edge applications.  
\paragraph*{5) Data Parallelism}
Data parallelism (also referred to as data splitting) is a technique that follows a decentralized architecture at the all in-edge level. A sizeable primary dataset is split in data parallelism to form mutually exclusive smaller datasets. These datasets are then forwarded to the processing ESs. In this architecture (see Figure \ref{fig501}),  the decision-making ES initially distributes the uninitialized model copy to each processing ESs. The processing ES starts training after it receives the dataset and the initial model copy. The decision-making ES is responsible for producing the global model by aggregating the local models residing inside the processing ESs.The global model is next sent back to the processing ESs so that it can continue to update its local model~\cite{negi2020distributed,li2020pytorch,szabo2020distributed}. 

\paragraph*{6) Federated learning }
\label{subsec_fedlearning}

\begin{figure}[h]
\centering 
\setlength
\fboxsep{0pt} 
\setlength\fboxrule{0.25pt} 
\fbox{\includegraphics[width=3.0in]{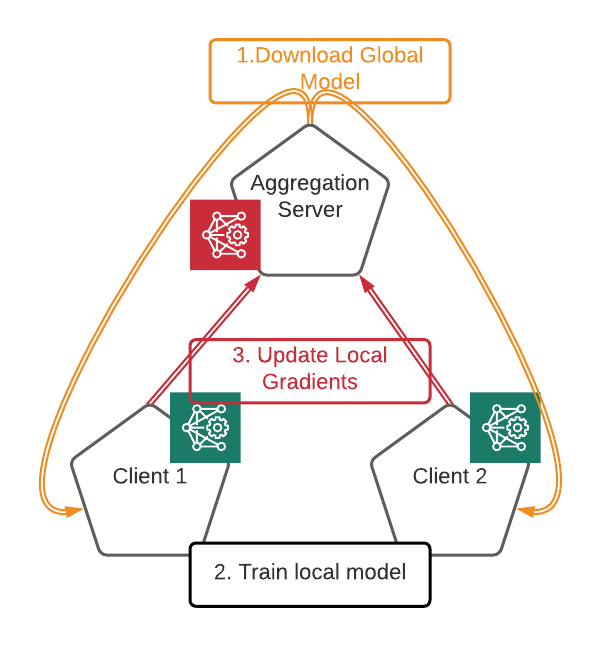}} \caption{Vanilla federated learning.} 
\label{fig7} 
\end{figure} 

Federated learning (FL) is a popular framework for training DL models using a decentralized and distributed architecture \cite{mcmahan2017communication}. Although the native framework treats mobile devices as clients responsible for training the DL model, recent research shows clients can be extended to the ES~\cite{khan2020federated,samarakoon2019distributed}, which makes this technology applicable for all in-edge. In this section, the `client' refers to processing ES with low computing resources, and the `aggregation ES' refers to decision-making ES with modestly higher computing capacity than the client. 

Federated learning enables ES to collaboratively learn a shared DL model while keeping all the training data on the client. As shown in Figure \ref{fig7}, during the first stage, all the clients download the global DL model from the aggregation ES, which is responsible for maintaining the global DL model. Once the global DL model is received, the client trains it using its own private data, making it a local DL model. Once training is completed on the client, the local model weights are sent to the aggregation ES. Once the aggregation ES receives all the weights from the participant client, it is then aggregated to formulate the new global DL model~\cite{bonawitz2019towards,reisizadeh2020fedpaq,long2022multi}. After aggregation, the global DL model is again circulated to the client for further training, making the whole approach cyclic. This framework ensures that the performance of the aggregated global model should be better than any of the individual client-side models \cite{li2021survey} before being disseminated.

Federated Learning Systems (FLS) can be further categorized based on their data partitioning strategy, privacy mechanism, and communication architecture~\cite{mothukuri2021survey,li2021survey,zhang2021survey,lyu2020threats}. The data partitioning strategy dictates how the data is partitioned across the clients. There are three broad categories of data partitioning (i) horizontal data partitioned FLS, (ii) vertical data partitioned FLS, and (iii) hybrid data partitioned FLS. In horizontal data partitioning, all the clients have the same attributes/features in their respective datasets needed to train the private DL model. Whereas in vertical data partitioned, all the clients have different attributes/features in the dataset. By utilizing entity alignment techniques (which helps find the overlap in other datasets where some of the features are common)~\cite{zhang2021asysqn,zhou2021privacy}, overlapped samples are collected for training machine learning models. 
\iffalse 
Wake-word recognition like 'Ok Google' is a good example of horizontal data partitioned FLS \cite{yu2021federated}, and when two different institutions come together, \emp{i.e.} bank and hospital to deduce about users will fall under vertical data partitioned FLS \cite{zhang2021survey}.
\fi
Hybrid data partitioning utilizes the best of both worlds. The entire dataset is divided into horizontal and vertical subsets in this category. So each subset can be seen as an independent dataset with fewer non-overlapping attributes and data points compared to the entire dataset~\cite{zhang2021survey}.\par
FLS provides privacy to a certain degree by default by allowing raw data to stay only with the client ES. However, while exchanging the model parameters, there is the possibility that exchanged model parameters could still leak some sensitive information about private data~\cite{yin2021comprehensive}. Therefore, privacy mechanisms have been employed for FLS. These mechanisms can be subdivided into either cryptographic techniques or differential privacy techniques. Cryptographic techniques require that both the client and aggregation ES operate on encrypted messages. Two of the most widely used privacy-preserving algorithms are homomorphic encryption \cite{stripelis2021secure,madi2021secure,zhang2022homomorphic,fang2021privacy} and multi-party computation~\cite{mondal2021poster,mou2021verifiable,sotthiwat2021partially}. On the other hand, differential privacy introduces random noise to either the data or the model parameters~\cite{girgis2021shuffled,zhang2021privacy,truex2020ldp,zhao2020local}. Although random noise is added to the data or model parameters, the algorithm provides statistical privacy guarantees while ensuring that the data or model parameters can still be used to facilitate effective global model development.\par
The communication architecture of an FLS can be broadly subdivided into two subcategories: distributed and decentralized architectures. In a decentralized architecture, the aggregation server is responsible for collecting and aggregating the local models from each client. It then sends the updated global model for retraining to each client. In this architecture, communication between the processing ES and decision-making ES can happen in synchronous \cite{chai2021fedat,zhang2021survey} as well as in asynchronous \cite{zhang2021secure,chai2021fedat,ma2021fedsa,wan2022privacy} manner. One of the significant risks in a decentralized architecture setting is that the decision-making ES may not treat each processing ES equally. That is, the decision-making ES may have a bias toward specific processing ES due to their higher participation during a training phase. A distributed architecture can mitigate the potential issues of bias. A distributed architecture in federated learning can be based on a P2P scheme (ex., gossiping scheme as described in Section \ref{lbl:gossip}), a blockchain-based system, or a graph-based system. In a distributed architecture, all the participating ESs are responsible for acting as processing and decision-making ES. Therefore, if a gossip scheme is implemented to achieve the decentralized FLS, all the models will randomly share the updates with their neighbors \cite{lo2021flra,chen2021bdfl}. In contrast, if a blockchain system is implemented, it leverages smart contracts (SC) to coordinate the DL training, model aggregation, and update tasks in FLS \cite{toyoda2020blockchain,nguyen2021federated,rahman2020secure,lu2020blockchain,li2020blockchain}. Lastly, if graph-based FLS is implemented, each client will utilize the graph neural network model with its neighbors to formulate the global models \cite{barbieri2021decentralized,he2022spreadgnn,xing2020decentralized,liu2021glint}.

FLS provides a much-needed way of enabling the DL model training and inference at the all in-edge paradigm. With an FLS, one can easily integrate multiple low-resource ESs to help train the DL model at the edge. Also, based on the resources available at the edge and the communication overhead of FLS, one gets the freedom to select either a distributed or decentralized architecture. 

\paragraph*{7) Split learning}
\label{subsec_splitlearning}

\begin{figure}[h]
\centering 
\setlength
\fboxsep{0pt} 
\setlength\fboxrule{0.25pt} 
\fbox{\includegraphics[width=3.0in]{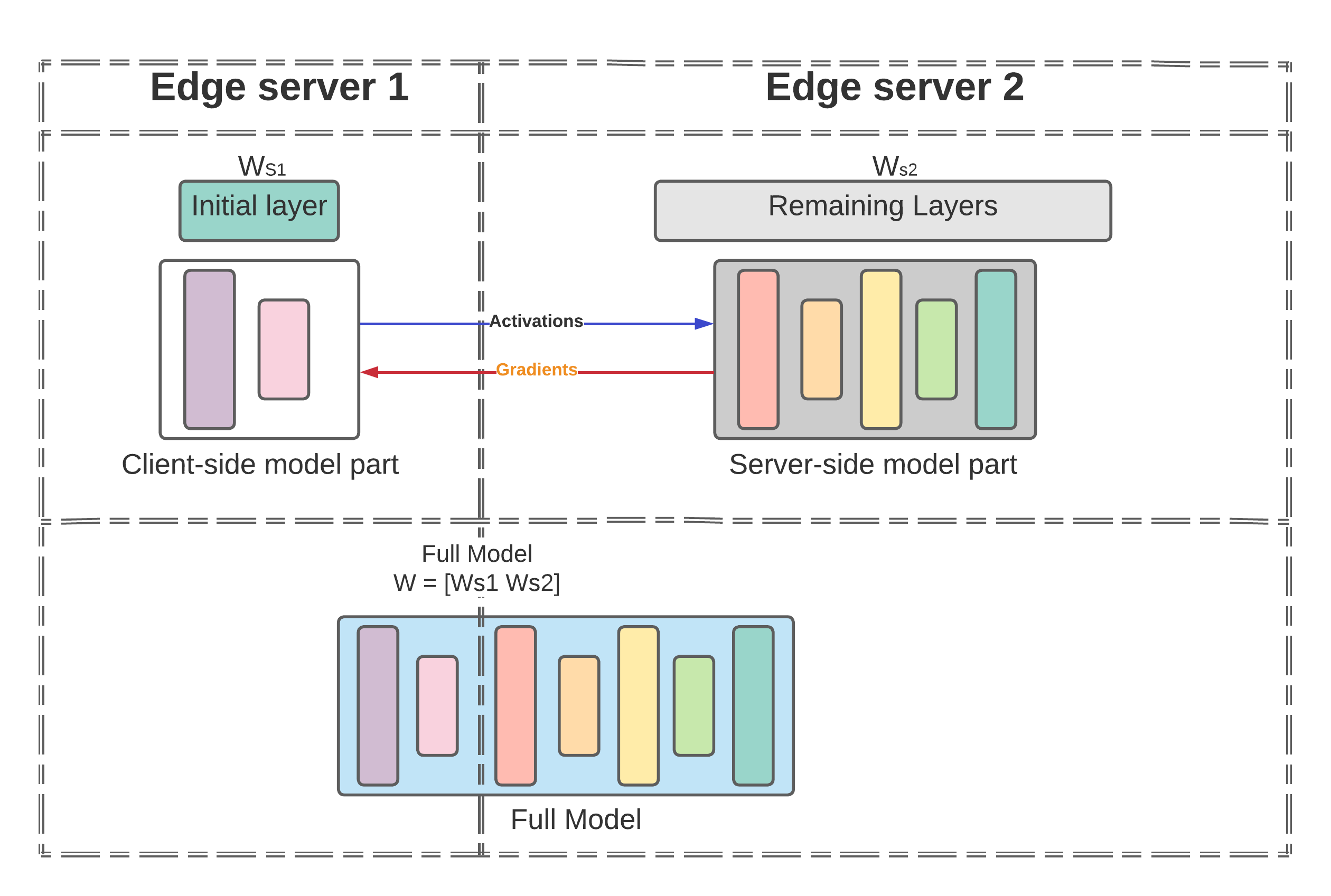}} \caption{Split Learning.} 
\label{SplitLearning_1} 
\end{figure}

In federated learning, each processing ES is responsible for locally training the whole neural network. In contrast, split learning provides a way to offload some of this computation between processing and decision-making ESs. More differences between federated learning and split learning are summarized in Table~\ref{tab:ComparisonOfet}. As we advance in this section, the `client' refers to processing ES with low computing resources, and the `server' refers to the decision-making ES with a modestly higher computing capacity than the client. Split learning divides a neural network into two or more sub-networks. Figure \ref{SplitLearning_1} illustrates the case where we split a seven-layer neural network into two sub-networks using layer 2 as the "cut layer". After the split, the two sub-networks are shared between the client, who trains the initial two layers of the network, and the server, who trains the last five layers of the network. At the training time, the client initiates the forward propagation of its confidential data and sends the activation from the cut layer to the server-side sub-network. The server then continues the forward propagation and calculates the loss. During backpropagation, gradients are computed and propagated initially in the server sub-network and then relayed back to the client side sub-network. In Split learning, during the training and testing, the server never gets access to the parameters of the client-side network or the client's data.

\begin{figure}[h]
\centering 
\setlength
\fboxsep{0pt} 
\setlength\fboxrule{0.25pt} 
\fbox{\includegraphics[width=3.0in]{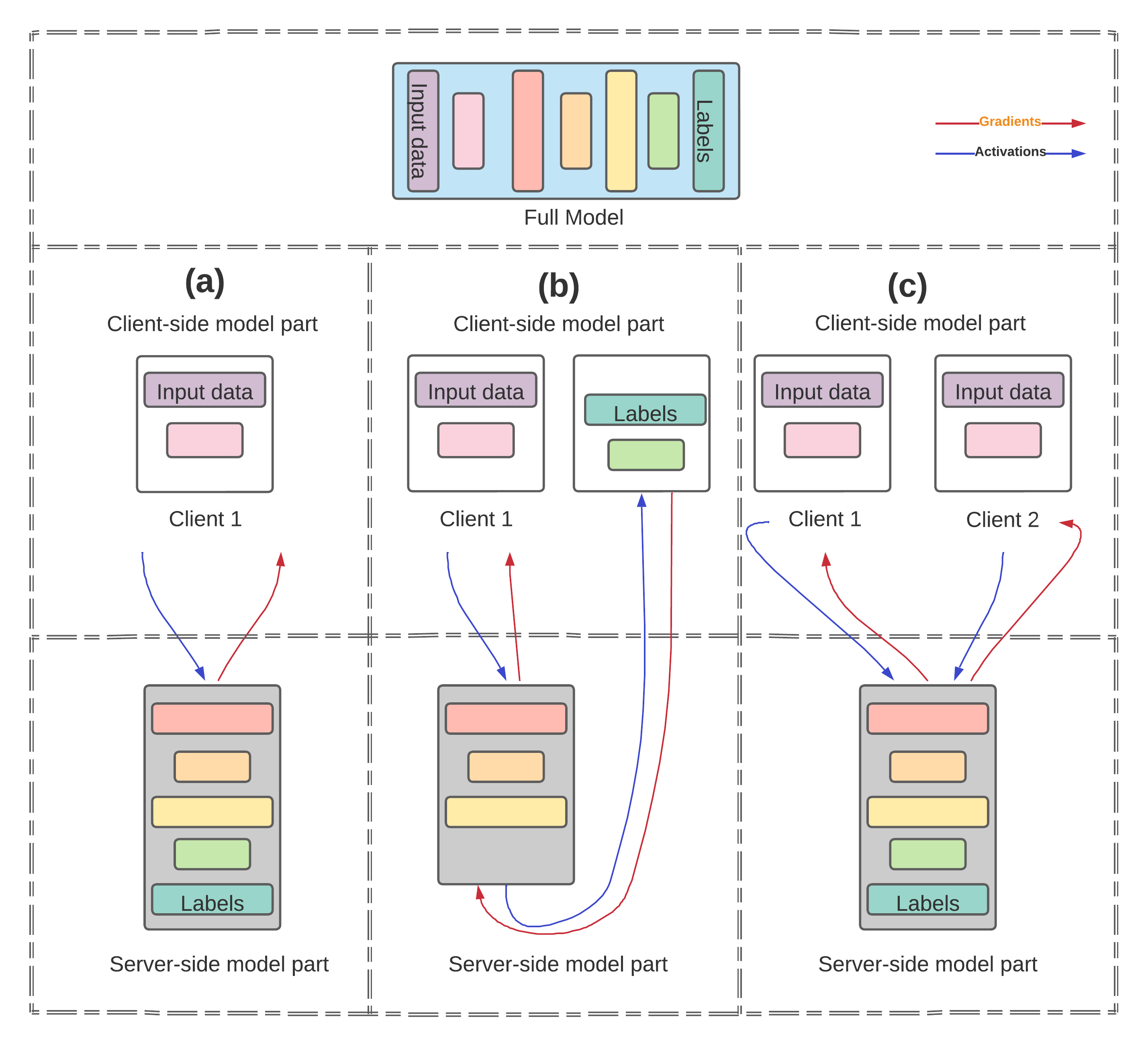}} \caption{Different configurations of Split Learning- (a)  simple vanilla split learning, (b) split learning without label sharing and (c) split learning for vertically partitioned data.} 
\label{Split Learning_Var} 
\end{figure}

Split learning can be broadly categorized into three configurations based on how the input data and labels are shared across the clients and servers. Figure  \ref{Split Learning_Var} shows three configurations- simple vanilla split learning, split learning without label sharing, and split learning for vertically partitioned data. A main neural network is partitioned into two sub-networks in simple vanilla split learning. The initial sub-network, along with the input data for the neural network, remains with the client, whereas the remaining sub-network, along with the labels, resides with the server \cite{thapa2021advancements}. Split learning without label sharing is identical to vanilla split learning, except that the labels reside with the client instead of the server. To compute the loss, the activations outputted from the server-side network are sent back to the client, who holds the last layer of neural network \cite{abuadbba2020can}. The loss is calculated, and gradients are computed from the last layer held by the client and then sent back to the server, and backpropagation takes place in the usual way. The final configuration of split learning is where the clients train their partial sub-network for vertically partitioned data and then propagate the activations to the server-side sub-network. The server-side sub-network then concatenates the activations and feeds them to the remaining sub-network. In this configuration, labels are also shared with the server \cite{vepakomma2018split}. \par

In a federated learning system, clients can interact with the server in parallel, which helps achieve faster training compared to a split learning approach. In split learning, the server must wait for all clients to send their activations before propagating the activation through the server-side network. Also, in contrast to federated learning, split learning reduces the computational requirements on the client-side (as only a partial amount of the network resides with the client). Recently, to leverage the advantages of both split learning and federated learning, a hybrid technique called splitfed learning was proposed \cite{thapa2020splitfed}. \par
\begin{figure*}[]
\centering 
\setlength
\fboxsep{0pt} 
\setlength\fboxrule{0.25pt} 
\fbox{\includegraphics[width=6.2in]{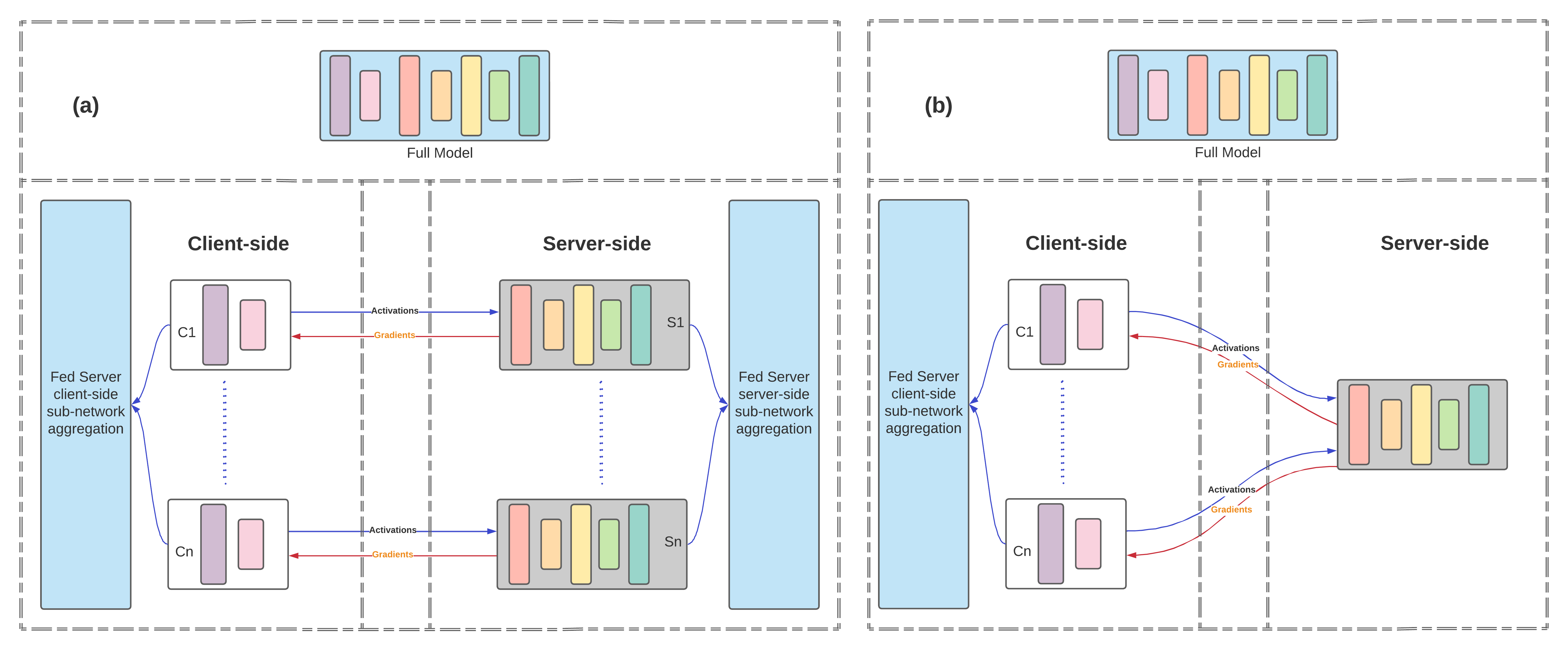}} \caption{Variants of splitfed learning (a) Splitfed learning with the same number of client and server-side sub-networks and (b) Splitfed learning with only one copy of server-side sub-network.} 
\label{SplitFed Learning} 
\end{figure*}

In splitfed learning, a DL model is broken down into the sub-networks shared amongst the clients and servers. In addition, there is a separate federated aggregation server for the client and the servers. All the clients perform the forward pass in parallel and independently of each other (as not seen in split learning). The resulting activations are sent to the server-side sub-network, which performs a forward pass for the remaining sub-network portion. The server then calculates the loss and back propagates the gradients back to the very first layer on the client-side, as described earlier with split learning. Once this process finishes, the servers send their model weights to a federated aggregation server, which aggregates the independent server-side sub-network to form a global server-side model. Similarly, the clients send their sub-network weights to another aggregation server. At the end of aggregation, a global model can be developed by combining the aggregated client-side weights with the aggregated server-side weights as shown in Figure \ref{SplitFed Learning} (a) \cite{thapa2020splitfed,pant2021comparison}. 

Splitfed learning can have several variants. For example, the first one is where each client has its own corresponding server-side network in the main server, \emph{i.e.}, the number of client-side models is equal to the number of server-side models as explained in the earlier paragraph. In the second variant, there are multiple clients but only a single server. Therefore, each client-side model sends its activations to a single common server-side sub-network, thereby reducing the required aggregation step and the need to keep multiple copies of the server-side networks as compared to the first variant as shown in Figure \ref{SplitFed Learning} (b). Moreover, as the server keeps only one copy of the server-side sub-network, it makes the server-side do forward and backward pass sequentially with each of the client's data (activations of the cut layer)~\cite{gawali2021comparison,joshi2022performance}.

\textbf{Key takeaways:}
The above-mentioned enabling technologies at the confluence of DL and all in-edge contribute to our understanding of training DL models using only ESs. The enabling technologies help address issues such as limited computational resources, communication overhead and latency between ESs, data privacy, and model robustness. Model parallelism and split learning provide a means of decreasing the computational resource required by individual ES. By splitting the DL model, multiple resource-constrained ESs can train a few layers of the network (rather than the entire model). Aggregated frequency control and federated
learning enable parallel model training, facilitating faster model convergence. Gossip training, federated learning, and aggregated frequency control adopt a distributed architecture, thereby robustly training a DL model in situations where the reliability of an ES is not predictable. Also, we have discussed splitfed, in which federated learning is combined with split learning. Splitfed overcomes the drawback of federated learning of training a large ML model in resource-constrained ESs~\cite{zhang2021fedsens}. At the same time, it eliminates the weakness of split learning to deal with one client at a time while training~\cite{joshi2022performance}.

\subsection{All in-edge model adaption}
\label{modelAdaption}
Model Adaption techniques provide a means by which DL model deployment at the ES can be achieved despite the lack of computing resources, storage, and bandwidth. Model adaption techniques can be broadly categorized into model compression and conditional computation techniques, as summarized in Table ~\ref{tab:modelAdaption}.

\paragraph*{1) Model Compression}
Model compression techniques facilitate the deployment of resource-hungry DL models into resource-constrained ES by reducing the number of parameters or training DL models that have been reduced in size from the original model. Model compression exploits the sparse nature of DL models by compressing the model parameters. Model compression reduces the computing,  storage, memory, and energy requirements needed for all in-edge deployment of DL models. This section reviews pruning, quantization, knowledge distillation, and low-rank factorization.

%\begin{enumerate}
\paragraph*{a) Pruning}
Pruning of parameters is the most widely adopted approach to model compression. This approach evaluates DL model parameters against their contribution to predicting the label. Those neurons that make a low contribution in inference are pruned from the trained model. Parameter pruning can significantly reduce the size of a DL model, but it also has the potential to impact the model’s performance adversely. In \cite{han2015learning}, the authors were able to reduce the size of the AlexNet and VGG-16 by a factor of $9\times$ and $13\times$ respectively, without incurring any loss in the accuracy over the ImageNet dataset. In another work~\cite{han2017ese}, the authors utilized pruning to create a compressed speech recognition model on field-programmable-gate-array (FPGA). This technique compressed the LSTM model by $10\times$ with negligible loss in accuracy. SS-Auto \cite{li2020ss} is a single-shot structured pruning framework.
In contrast to earlier versions of pruning where the entire DL model's parameters were selected for pruning, in structured pruning, independent pruning on columns and rows of filters and channels matrix (for CNN-based DL models) is performed. The compressed DL models produced by the SS-Auto framework did not suffer any degradation in performance, achieving the original performance levels when tested on CIFAR-10 and CIFAR-100 datasets. However, the compressed VGG-16 model reduced the number of convolutional layers parameters by a factor of 41.4\% for CIFAR-10 and 17.5\% for the CIFAR-100 dataset. In \cite{gong2020privacy}, the authors proposed a new framework based on weight pruning and compiler optimization for faster inference while preserving the privacy of the training dataset. This approach initially trains the DL models as usual on the user’s data. The model then undergoes privacy-preserving-oriented DNN pruning. Finally, synthetically generated data (with no relevance to the training data) is passed through a layer of the user-trained model. 

\begin{landscape}
\begin{table}[]
\centering
\caption{Comparison of enabling techniques for all in-edge.}
\label{tab:ComparisonOfet}
\begin{tabular}{|p{0.1\textwidth}|p{0.2\textwidth}|p{0.1\textwidth}|p{0.1\textwidth}|p{0.1\textwidth}|p{0.1\textwidth}|p{0.1\textwidth}|p{0.1\textwidth}|} 
\hline\hline
\multicolumn{1}{|c|}{Category}       & \multicolumn{1}{c|}{ Model Parallelism/ DNN Splitting}            & \multicolumn{1}{c|}{Aggregation Frequency Control
}              & \multicolumn{1}{c|}{Gossip Training}               & \multicolumn{1}{c|}{Gradient Compression}               & \multicolumn{1}{c|}{Data Parallelism}               & \multicolumn{1}{c|}{Federated Learning}               & \multicolumn{1}{c|}{Split Learning}     \\ 
\hline \hline
Partial model training      &    Yes
& No
& No
& No
& No
& No
& Yes    \\ 
\hline
Parallel model training      &   No
& Yes
& Yes
& Applicable for parallel/ non-parallel model training.
& Yes
& Yes
& No \\ 
\hline
Model      &    Single DNN model is partitioned across multiple ESs.
& Multiple clusters of ESs collaboratively train the model.
& Multiple ESs train model collaboratively.
& Helps to reduce the size of gradients passed across ESs during collaborative training.  
& Large dataset is split and passed to multiple ESs to train the model collaboratively.
& Clients server collaboratively trains the model.
& Client server shared model architecture and collaborative training.                   \\ 
\hline
System Architecture       &     Decentralized
& Decentralized
& Distributed
& Decentralized/ Distributed
& Decentralized
& Decentralized/ Distributed
& Decentralized/ Distributed       \\ 
\hline
Inter-communication during model training      &    Only activation vectors from the last layers are shared.
& Full model parameters exchanged.
& Full model parameters exchanged.
& Compressed model parameters exchanged.
& Full model parameters exchanged.
& Full model parameters exchanged.
& No model parameters are exchanged (only activation vectors from the last layers are shared)  \\ 
\hline
Computational resource requirements for large DNN      &    Equal computation resources are required at each Ess.
& High at client and server end.
& Equal computation resources are required at each ESs.
& Equal computation resources are required at each ESs.
& Equal computation resources are required at each ESs.
& High at client and server end.
& Low at the client end and comparatively high at the server end.
  \\ 
\hline
Data privacy by default      &    No
& Yes
& Yes
& No
& No
& Yes
& Yes\\ 
\hline
Communication overhead between server and clients      &  Depends on a number of partitions.
& Depends on model size.
& Depends on model size.
& Reduced
& Depends on the number of partitions.
& Depends on model size.
& Depends on sample size and the number of nodes in the cut-layer.  \\
\hline
Related Research      &  \cite{kim2016strads,narayanan2021efficient,geng2019elasticpipe,mao2018privacy,xu2020acceleration,yoon2021edgepipe}      &  \cite{hsieh2017gaia}     &  \cite{hegedHus2021decentralized,dinani2021gossip,kong2021consensus,nikolaidis2021using}      &  \cite{tang2018communication,strom2015scalable,tao2018esgd,abrahamyan2021learned,chen2020scalecom}        &   \cite{li2015malt,szabo2020distributed,yu2021toward,park2020hetpipe,li2020distributed,cheikh2021klessydra}     & \cite{bonawitz2019towards,reisizadeh2020fedpaq,long2022multi,mothukuri2021survey,li2021survey,zhang2021survey,lyu2020threats,chen2022evfl,jin2021cafe,zhou2021privacy}         & \cite{abuadbba2020can,thapa2022splitfed,thapa2021advancements,thapa2020splitfed,pant2021comparison}    \\
\hline \hline
\end{tabular}
\end{table}
\end{landscape}

The decision to prune a parameter or not from the current layer is based on how similar (by computing the Frobenius norm) the original output of the layer (without pruning) is when compared with the output of the layer after the parameter has been pruned. If the outputs are close enough, then that parameter is pruned. This pruning technique is named the alternating direction method of multipliers (ADMM). Experimental results of the framework outperformed the state-of-the-art end-to-end frameworks, i.e., TensorFlow-Lite, TVM, and MNN, with speedup in inference up to $4.2\times$, $2.5\times$, and $2.0\times$, respectively.

\paragraph*{b) Quantization}

Data quantization degrades the precision of the parameters and gradients of the DL model. More specifically, in quantization, data is represented in a more compact format (lower precision form). For example, instead of adopting a 32-bit floating-point format, a quantization approach might utilize a more compact format such as 16-bit to represent layer inputs, weights, or both \cite{zhou2019EDGE}. Quantization reduces the memory footprint of a DL model and its energy requirements. In contrast, pruning the neurons in a DL model will reduce the network's memory footprint but does not necessarily reduce energy requirements. For example, if later-stage neurons are pruned in a convolutional network, this will not have a high impact on energy because the initial convolutional layer dominates energy requirement \cite{zhou2019EDGE}.
In \cite{yang2021dynamic}, the authors utilized a dynamic programming-based algorithm in collaboration with parameter quantization. With the proposed dynamic programming-assisted quantization approach, the authors demonstrated a $16\times$ compression in a ResNet-18 model with less than a 3\% accuracy drop. The authors in \cite{huang2021mixed} proposed a quantization scheme for the inference phase of the DL model that targets weights along with the inputs to the model and the partial sums occurring inside the hardware accelerator. Experiments showed that the proposed schema reduced the inference latency and energy consumption by up to $3.89\times$ and $4.84\times$, respectively, while experiencing a 1.18\% loss in the DL models inference accuracy. 

\begin{figure}[ht]
\centering 
\setlength
\fboxsep{0pt} 
\setlength\fboxrule{0.25pt} 
\fbox{\includegraphics[width=3.0in]{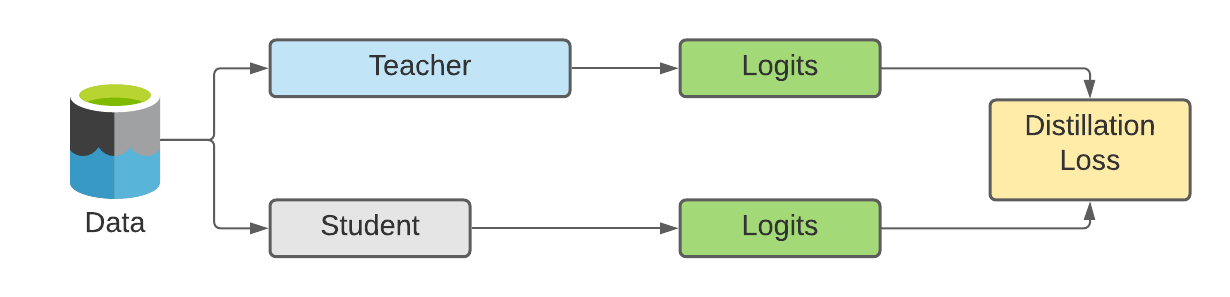}} \caption{Teacher-student architecture for Knowledge Distillation.} 
\label{fig_KD} 
\end{figure}

\paragraph*{c) Knowledge Distillation}
Knowledge distillation is a  model compression technique that helps train a smaller DL model from a significantly larger trained DL model. The knowledge distillation comprises three key components: (i) The original knowledge, (ii) the distillation algorithm, and (iii) the teacher-student architecture \cite{gou2021Knowledge}. The original knowledge is the original large DL model, which is referred to as the teacher model. The knowledge distillation algorithm
is used to transfer knowledge from the teacher model to the smaller student model using techniques such as Adversarial KD
\cite{mirzadeh2020improved,tsunashima2021adversarial}, Multi-Teacher KD \cite{yuan2021reinforced,wang2021mulde,hao2021model}, Cross-modal KD \cite{thoker2019cross,sun2021unsupervised}, Attention-based KD \cite{passban2021alp,inaguma2021alignment,you2021contextualized,chen2021cross}, Lifelong KD \cite{chuang2020lifelong,yao2020Knowledge} and Quantized KD \cite{shen2020q,boo2021stochastic}. Finally, the teacher-student architecture is used to train the student model. A general teacher-student framework for Knowledge distillation is shown in Figure  \ref{fig_KD}. In this architecture, the teacher DL model is trained on the given dataset in the initial phase. Once the teacher DL model is trained, it assists the shallower student DL model. The student DL model also uses the same dataset used to train the teacher DL model, but labels for the data points are generated by the teacher DL model \cite{meng2019conditional}. The knowledge distillation technique helps a smaller DL model imitate the larger DL model's behavior.

KD provides a viable mechanism of model compression \cite{gou2021Knowledge}. This technique helps reduce the number of ESs required to deploy the larger DL model at the all in-edge level. Reduction in the number of ES also helps achieve faster inference time from ESs (as less communication needs to be done within ESs).

\iffalse
However, a mismatch in the accuracy during model evaluation indicates students' incapability to mimic the teacher perfectly \cite{cho2019efficacy,zhang2018deep,gou2021Knowledge}, which requires more research in the future.
KD provides a viable mechanism of model compression \cite{gou2021Knowledge,huang2017like}. However, a number of open research challenges still exist in the area:
\begin{enumerate}
    \item The teacher models can guide the student model on the task ({\em i.e.}, classification, regression etc.) in hand. Still, the student model is not able to learn all the significant Knowledge. The mismatch in the accuracy during model evaluation indicates students' incapability to mimic the teacher. Thereby, optimization methods are crucial for significant Knowledge absorption from the teacher to the student and require more research \cite{cho2019efficacy,zhang2018deep,gou2021Knowledge}.
    \item The student models are not able to follow the teacher models. This can happen when the model parameters of teacher models are significantly different from each other. On the one hand, the teacher models can be very deep in contrast to the shallower network of the student model. On the other hand, it is also seen if the student model is similar to the teacher model architecture, it will produce outputs identical to the teacher \cite{heo2019comprehensive,sun2020mobilebert}.
\end{enumerate}
\fi

\paragraph*{d) Low-Rank Factorization}
Low-rank factorization is a technique that helps in condensing the dense parameter weights of a DL model \cite{jain2021low,papadimitriou2021data}, limiting the number of computations done in convolutional layers \cite{,patrona2021self,han2021learning,lee2021training} or both \cite{yang2021iterative,russo2021dnn}. This technique is based on the concept of creating another low-rank matrix that can approximate the dense metrics of the parameter of a DL model, convolutional kernels, or both. Low-rank factorization can save memory on an ES while decreasing computational latency because of the resulting compact size of the DL model. In \cite{chen2022fpc}, the authors used the low-rank factorization by applying a singular value decomposition (SVD) method. They demonstrated a substantive reduction in the number of parameters in convolutional kernels, which helped reduce floating-point operations(FLOPs) by 65.62\% in VGG-16 while also increasing accuracy by 0.25\% when applied to the CIFAR-10 dataset. Unlike pruning, which necessitates retraining the DL model, after applying low-rank factorization, there is no need to retrain the DL model.
Further research \cite{swaminathan2020sparse} proposed a sparse low-rank approach to obtain the low-rank approximation. The sparse low-rank approach is based on the idea that all the neurons in a layer have different contributions to the performance of the DL model. So based on the neuron ranking (based on the contribution made for inference), entries in the decomposition matrix were made. This approach, when applied over the CIFAR-10 dataset with VGG-16 architecture, achieved $3.6\times$ times smaller compression ratio to the SVD. Other commonly used methods for low-rank factorization are tucker decomposition (TD) \cite{shi2021low,fu2022low,ma2021fast} and canonical polyadic decomposition (CPD) \cite{phan2021canonical,chantal2021dynamic}.

%\end{enumerate}

\paragraph*{2) Conditional Computation}

Conditional computational approaches alleviate the tension between the resource-hungry DL model and the resource-constrained ES. In conditional computation, the computational load of the DL model deployed over a single ES is distributed with other ES in the network. The selection of an appropriate conditional computation technique is based on the DL model’s latency, memory, and energy requirements. Therefore, depending upon the configuration of the ES and DL model's computation requirements, DL model deployment can utilize one or any combination of the techniques (Early Exit, Model Selection, and Result Cache) defined in this section.

%\begin{enumerate}
\paragraph*{a) Early Exit}
\label{lbl_early_exit}

\begin{figure}[ht]
\centering 
\setlength
\fboxsep{0pt} 
\setlength\fboxrule{0.25pt} 
\fbox{\includegraphics[width=3.0in]{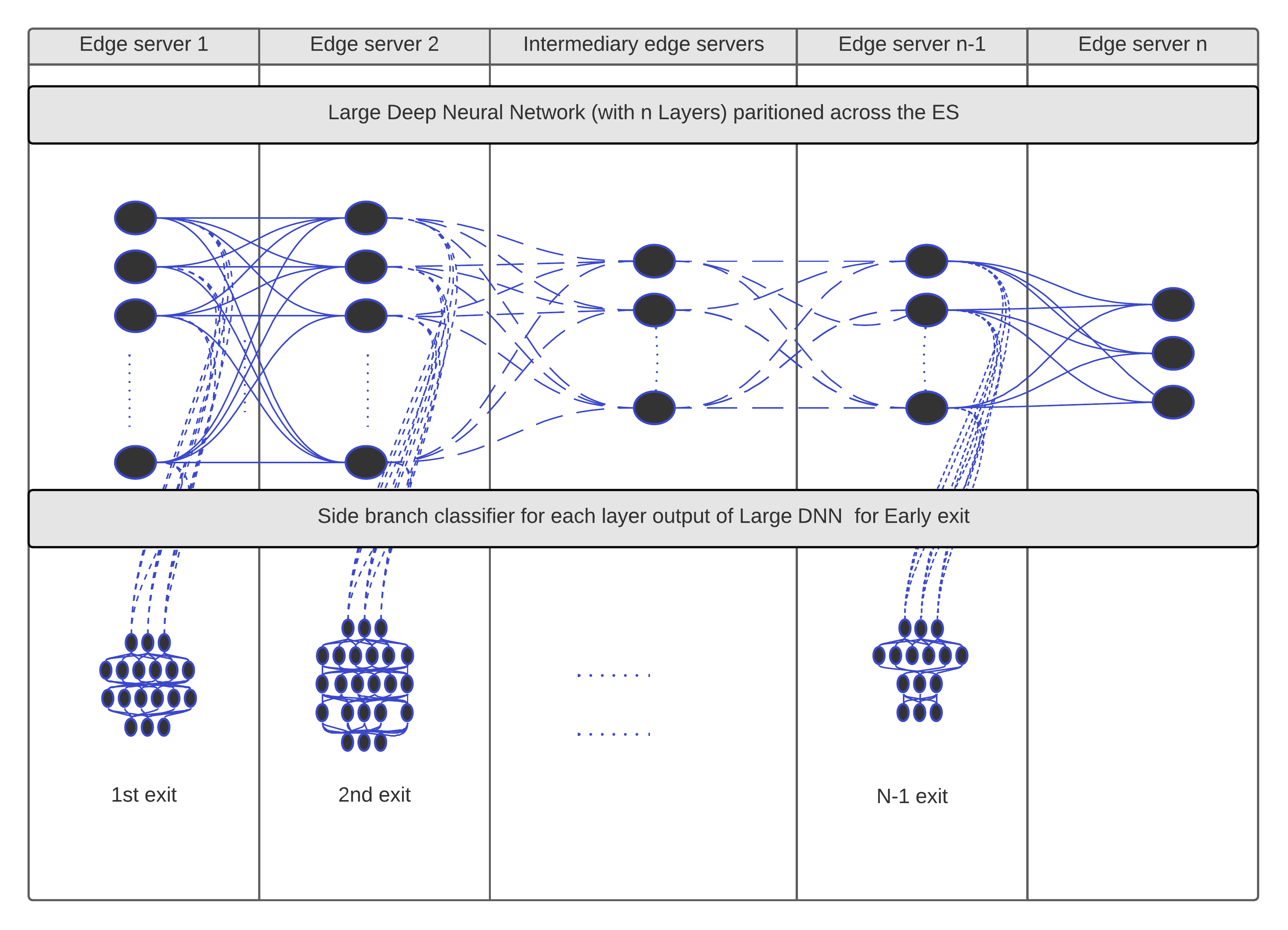}} \caption{Early exit adaption of Deep Neural Network.} 
\label{fig10} 
\end{figure}

The main idea behind the early exit approach is to find the best tradeoff between the deep DNN structure of a DL model and the latency requirements for inference. In this approach, a deep neural network trained on a specific task is partitioned across multiple ESs. The partitioning of the DL model is based on a layer-wise split, such that a single or multiple layers can reside across multiple ESs based on the computation power provided by each ES. Each ES that hosts one or more layers of the DL model also attaches a shallower model (or side branch classifier) to the output of the final layer on the current ES. The model is then trained as shown in Figure \ref{fig10}. The purpose of the side branch classifier is to provide an early prediction or early exit. During inference, the data is propagated through the network (and each ES host). Each host will calculate both the output of the hosted layers and the output of the local early exit network. If the output of the early exit layer exceeds a defined confidence threshold, then the propagation stops (this is the early exit), and the `early' result is returned. If the prediction from the early exit network is less than the confidence threshold, the output of the larger DL model's layers is then propagated to the next ES in the chain, which holds the next layer of the larger DL model and another early exit network. The process of propagating the layer’s output to the subsequent layer is carried out until one ES inferences the class with a higher confidence score. This process can provide $‘n - 1’$ exit points for a DL model with $‘n’$ neural network layers; thus, if layer 1 of the larger DNN along with the side branch can infer the class with the required confidence that output will be given as a response to the end user eliminating any further propagation of activation values along the ES.

Researchers in \cite{teerapittayanon2016branchynet} provided the programming framework\newline `Branchynet’, which helps incorporate the early exit approach into a standard DL model. The framework modifies the proposed DL model by adding exit branches at certain layers. With the multiple early exit points, it can also be considered as an enabler for localized inference using DL models with less number of layers. For the AlexNet DL model, ‘Branchynet’ framework was able to reduce the inference time by a factor of $2\times$ and $6\times$ on CPU and GPU, respectively. In \cite{li2021deepqtmt}, the authors proposed DeepQTMT to lower the encoding time spent on video compression. In the DeepQTMT, the authors utilized a multi-stage early exit mechanism to reduce the high encoding time. Experimental results showed the encoding time was reduced by a factor ranging from 44.65\% to 66.88\% with a negligible adverse impact in bit-rate within the range of 1.32\% to 3.18\%. Therefore, while the early exit strategy can decrease latency and facilitate a faster response time, it does have the drawback of increasing the memory footprint of the DL model, thus utilizing more storage at each ES. 
% \iffalse
% But due to increases in the branches' memory footprint of the DNN increases significantly compared to the vanilla version of the same DNN. An early exit in the all in-edge paradigm helps utilize the ES, which are very close to the source of data generation. If inference does not hold good, it then passes the activation to the subsequent ES, which is far from the point of data generation. Although this technique helps lower the latency to provide faster inference, it utilizes more storage at individual ES.
% \fi

\paragraph*{b) Model Selection}

The model selection approach selects a specific DL model for inference from a set of available DL models based on the latency, precision, and energy requirements of the end user \cite{wang2020convergence}.  In a model selection strategy, multiple DL models with varying DL model structures are trained. The different trained models each have a specific inference latency, energy requirements, and accuracy. Once trained, each of the models is deployed to various servers. The model selection approach will then select the DL model based on the end user requirements  \cite{zhou2019EDGE}. \par
\begin{figure}[ht]
\centering 
\setlength
\fboxsep{0pt} 
\setlength\fboxrule{0.25pt} 
\fbox{\includegraphics[width=3.0in]{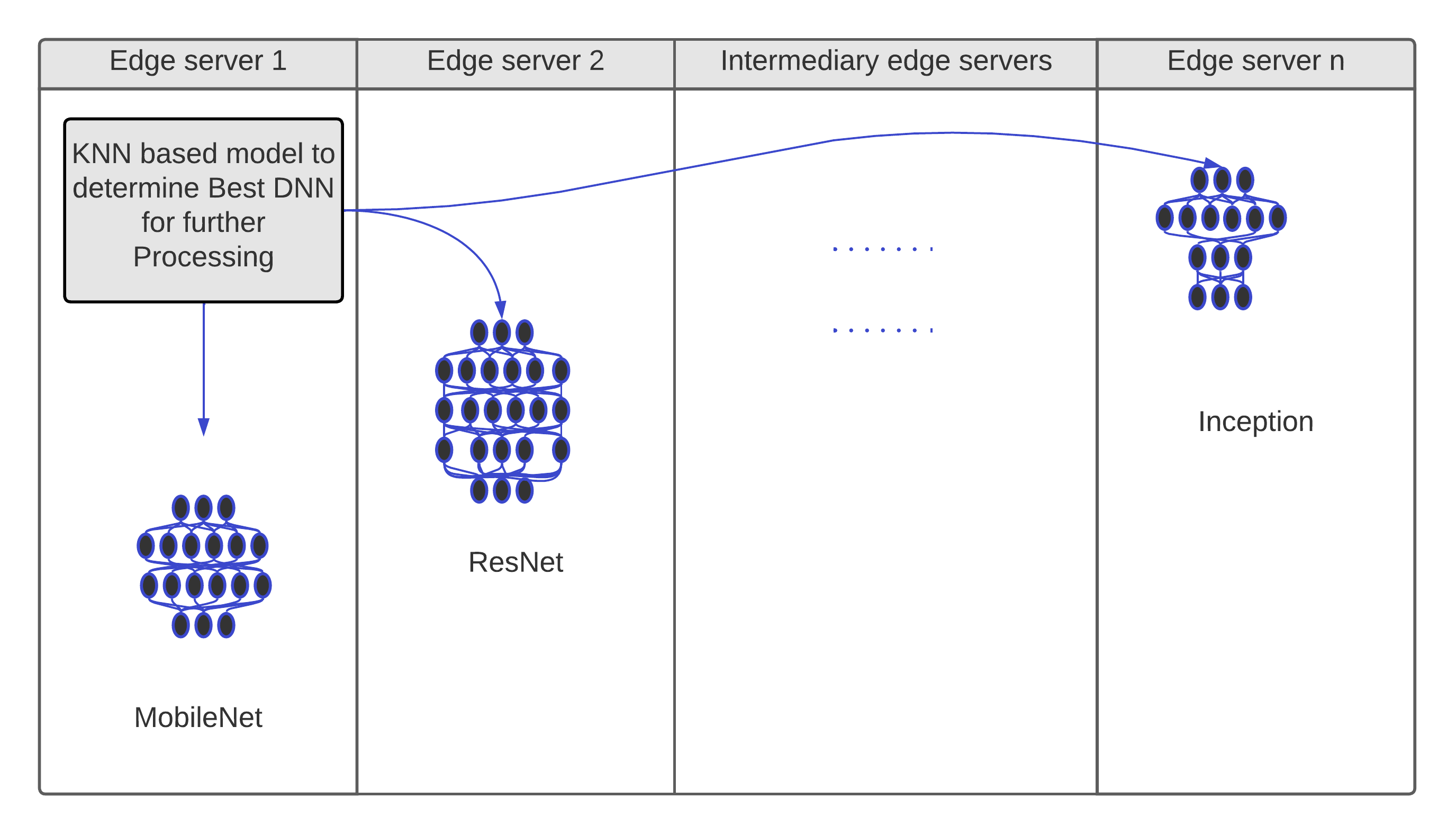}} \caption{Model Selection of Deep Neural Network.} 
\label{fig:modelselection} 
\end{figure}

The model selection approach is similar to the early exit approach, with only one difference. In model selection, independent DL models are trained; in contrast, in the early exit, only one DL model is trained over which multiple exit points are created. Authors in \cite{park2015big} proposed a new concept of BL-DL (big/little DL) based on the model selection approach. The authors proposed the score margin function, which helps in deciding whether or not the inference made by a small DL model is valid. The score function is computed by subtracting the highest probability from the second-highest probability of a class from the last classifier layer of a DL model.  Thus, a score function can be seen ranging from 0 to 1. The higher the value of the score function, the higher the estimation that inference is accurate. The lower the value of the score function, the lower the estimation of inference being accurate. If the score function estimation is low, then a larger DL model is invoked to make the inference on the same input data. The same research showed a 94.1\% reduction in the energy consumption on the MNIST dataset, with accuracy dropping by 0.12\%. Recently in \cite{marco2020optimizing}, an adaptive model selection technique has been used to optimize the DL model's inference. The proposed framework builds a standard DL model, which learns to predict the best DL model to use for inference based on the input feature data. To facilitate the training of the selection model (which is the standard KNN model in this scenario), different pre-trained models like Inception \cite{liu2021deep}, ResNet \cite{sarwinda2021deep}, MobileNet \cite{kadam2021detection} were evaluated on the same image dataset. For each image, the DL model that achieved the highest accuracy is set as the output. The training data for the KNN model comprises the features extracted from the image as input and the optimal DL model as output. Once the model selector (the KNN) is trained, it is then used to determine the DL model, giving the best accuracy on the selected image. In the end, the selected DL model makes an inference on the image as shown in Figure~\ref{fig:modelselection}.

Experimental results validated the reduction in the inference time by a factor of $1.8\times$ for the classification task and $1.34\times$ time reduction in a machine translation task. While model selection facilitates a decrease in inference time, it does incur an increased memory footprint across the ESs due to the number of pre-trained DL models.  

\paragraph*{c) Result Cache}
\label{lbl_result_cache}

Result cache techniques help in decreasing the time required to obtain the prediction from the ES. In this approach, frequent input queries (such as frames in the case of video classification or images in the case of image classification) and the associated predictions made by the DL model are saved in an archive on the ES. So, before any query is inferred from the DL model, intermittent lookup happens. In intermittent lookup, if a query is similar to a saved query, the result is inferred from the archive (cache). Otherwise, the query goes to the DL model for inference. This technique becomes more powerful in environments where the queries can be expected to exhibit similarity.
In \cite{drolia2017cachier}, the authors proposed a cache-based system that leveraged the ES for image classification. When evaluated on image classification applications, the approach yielded up to $3\times$ speedup on inference for image recognition tasks without any drop in the model’s performance (accuracy). Another system for video analysis utilized the cached convolution outputs of the CNN layers to reduce the computation for making an inference \cite{huynh2017deepmon}. The idea is again based on the similarity of consecutive frames in videos. Initially, in this approach, activations from each layer of DL for a query frame are saved in the cache. For the next subsequent frame (query), the query is pushed through the first layer, and the resulting activations are compared with the previous activation values of the same layer saved in the cache. Only those activations that differ significantly from the cached version are calculated and propagated further through the network. If the activation is deemed similar, they are carried over with their cache results to the next layer. In the experiment, the authors showed a significant speedup of $3\times$ to $4\times$ compared to the vanilla CNN model with no change in accuracy. In other research \cite{balasubramanian2021accelerating}, the authors proposed a framework similar to result caching. In this research, queries were initially passed through the DL model, and activations of each layer were cached (archived) in the ES along with the prediction from the DL model. During the inference, after passing the image through the layers of DL, activations are checked with the saved activations of a specific layer. If the activations of a particular layer for the current query match with the activations in the cache, further propagation of the activations is stopped, and the cached result is returned as the prediction. This research was applied to a VGG-16 architecture using CIFAR and yielded a $1.96\times$ latency gain using a CPU and a $1.54\times$ increase when using a GPU with no loss in accuracy. Result caching provides a significant boost in the scenario where the query (frames processing for the boundary identification) for inference does not change significantly. While result caching improves the overall latency of the neural network, it also incurs a larger memory footprint. 
\newline

\begin{landscape}
\begin{table}[]
\centering
\caption{Comparison of model adaption techniques for all in-edge.}
\label{tab:modelAdaption}
\begin{tabular}{|p{0.1\textwidth}|p{0.2\textwidth}|p{0.1\textwidth}|p{0.1\textwidth}|p{0.1\textwidth}|p{0.1\textwidth}|p{0.1\textwidth}|p{0.1\textwidth}|} 
\hline\hline
\multicolumn{1}{|c|}{Category}       & \multicolumn{1}{c|}{Pruning}            & \multicolumn{1}{c|}{Quantization}              & \multicolumn{1}{c|}{Knowledge distillation}               & \multicolumn{1}{c|}{Low rank factorization}               & \multicolumn{1}{c|}{Early exit}               & \multicolumn{1}{c|}{Model selection}               & \multicolumn{1}{c|}{Result cache}     \\ 
\hline \hline
Model adaption category
& Model compression
& Model compression
& Model compression
& Model compression
& Conditional computation
& Conditional computation
& Conditional computation    \\ 
\hline
Number of DL models involved
& Single
& Single
& Single
& Single
& Multiple
& Multiple
& Single/ Multiple    \\ 
\hline
DL computation
& Reduced
& Reduced
& Reduced
& Reduced
& It can be reduced or increased based on exit criteria.
& Depends on the DL model selected for making inferences.
& No impact\\ 
\hline

DL model size
& Reduced
& Reduced
& Reduced
& Reduced
& Increased
& Depends on DL model selection technique.
& No impact\\ 
\hline

Accuracy drop due to model adaption technique
& Yes
& Yes
& Yes
& No
& No
& No
& No\\ 
\hline

Memory footprint
& Reduced
& Reduced
& Reduced
& Reduced
& Increased
& Increased
& Increased\\ 
\hline

Inference time
& Reduced
& Reduced
& Reduced
& Reduced
& It can be reduced or increased based on exit criteria.
& Depends on the DL model selected for making inferences.
& Reduced\\ 
\hline
Related Research      &  \cite{qiu2020pre,berthelier2021deep,liang2021pruning,xu2020convolutional,gao2020rethinking,han2015learning,han2017ese,li2020ss,gong2020privacy}     &  \cite{liang2021pruning,berthelier2021deep,naik2021survey,zhang2021making,zhou2019EDGE,yang2021dynamic,huang2021mixed}    &  \cite{wang2018not,sharma2018existing,niu2021distant,hazarika2021conversational,soleimani2021cross,chuang2020lifelong,yao2020Knowledge,passban2021alp,inaguma2021alignment,chen2021cross,sun2021unsupervised,yuan2021reinforced,wang2021mulde,hao2021model,mirzadeh2020improved,tsunashima2021adversarial}   &  \cite{jain2021low,papadimitriou2021data,patrona2021self,han2021learning,lee2021training,yang2021iterative,russo2021dnn}  &  \cite{baccarelli2021learning,tan2020end,passalis2020efficient,tan2021empowering,laskaridis2020hapi,teerapittayanon2016branchynet,wang2020convergence,li2021deepqtmt}  &  \cite{park2015big,zhou2019EDGE,marco2020optimizing,wang2020convergence}      &  \cite{kumar2020quiver,krichevsky2021quantifying,romero2021memory,drolia2017cachier,huynh2017deepmon,balasubramanian2021accelerating,cheng2020adaptive,zong2020efficient,wang2021diesel+,inci2020deepnvm}  \\
\hline \hline
\end{tabular}
\end{table}
\end{landscape}

\textbf{Key takeaways:}
This section described model adaption techniques, which facilitate the efficient deployment of large DL models at the all in-edge level, divided into segments- model compression and conditional computation, summarized in Table \ref{tab:modelAdaption}. \par
Model compression techniques such as pruning, quantization, knowledge distillation, and low-rank factorization provide practical ways of reducing the size and memory footprint of the DL model. Reduction in model size due to model compression techniques also decreases the amount of computation needed for making an inference. However, there lies a need for DL model retraining while adopting pruning and knowledge distillation, whereas no retraining is required for quantization and low-rank factorization. Also, a drop in accuracy is observed amongst all model compression techniques leaving low-rank factorization. \par
Conditional computation techniques such as early exit, model selection, and result caching provide practical ways to utilize the computational resources of the available ESs to provide faster inference. In contrast to a reduced memory footprint observed while using the model compression technique, the memory footprint increases while adopting conditional computation. Also, no significant drop in accuracy is observed while utilizing the conditional computation techniques.
%===================================================================
\section{Key Performance Metrics of all in-edge}
\label{sec:util}
The application of DL at the edge has gathered significant momentum over the last few years. 
Typically research evaluates the performance of a limited number of DL models often adopting a different set of standard performance metrics (such as top-k accuracy~\cite{khoa2021fed} and mean average precision~\cite{yang2022distilled}). Unfortunately, these standard metrics fail to provide insights into the runtime performance of DL model inference at ESs. Relevant performance metrics for DL services include but are not limited to latency, use-case-specific metrics, training loss, communication cost,  privacy-preserving metrics, energy consumption, memory footprint, combined metrics, robustness, transferability, and lifelong learning.  Table~\ref{tab:KPI}, summarizes the metrics/ description against KPI utilized at all in-edge.  \par
This section will discuss the different metrics that should be evaluated when developing all in-edge based DL models.

\begin{table*}[]
\centering
\caption{Key performance metrics at all in-edge}
\label{tab:KPI}
\begin{tabular}{|p{0.3\textwidth}|p{0.6\textwidth}|}
\hline\hline
\multicolumn{1}{|c|}{\textbf{Performance Indicator}} & \multicolumn{1}{c|}{\textbf{Metrics/ Descriptions}} \\ \hline \hline
Use-case specific metrics & Accuracy, F1-score, Precision, Recall, R-square, Root mean squared error (RMSE), etc. \\ \hline
Training loss & Mean absolute error, Mean square error, Negative log-likelihood, Cross-entropy, etc. \\ \hline
Convergence rate & The number of iterations taken by a set of DL models (either having different model hyper-parameters or architecture) to converge on the same solution.\\ \hline 
Latency   & Computational latency and Communication latency          \\ \hline 
Communication cost    & The amount of data (message size of each query) flowing to the deployed DL model at an ES from the end user. Communication cost is measured in kilobytes (KB) or megabytes (MB).\\ \hline
Privacy-preserving &  Mutual information score (MIS),  Structural similarity index measure (SSIM), Matthews correlation coefficient, etc.  \\ \hline
Energy consumption &  The amount of energy consumed by a set of DL models at an ES during the training and inference phase. Energy consumption is measured in watts and kilowatts.  \\ \hline
Memory footprint & The amount of memory space (in RAM) required to
host a set of DL models at ES during the training and inference phase. Model size or memory footprint is measured in megabytes (MB).   \\ \hline
Combined metrics & Energy-precision ratio (EPR).  \\ \hline
Robustness & KL Divergence \\ \hline
Transferability and lifelong learning &   Transfer accuracy and Log expected empirical prediction (LEEP) score \\ \hline \hline
\end{tabular}
\end{table*}

\subsection{Use-case specific metrics}
Use-case specific metrics are used to determine the quality of the trained DL model and are dependent on the problem statement. For example, if the use-case is a classification problem, then accuracy, F1-score, roc\_auc, etc. can be evaluated \cite{kristiani2020isec,singh2021deep,yu2021deep}. 
Accuracy and F1-score are the most common metrics used to determine the quality of classification problems. In the classification problem, the DL model is trained to correctly predict the class of interest, i.e., true positive (TP) and the class of dis-interest, i.e., true negative (TN). Equation~\ref{m_acc} represents the mathematical formulation of Accuracy, where FP is a false positive (classes that are wrongly classified as positive), and FN is a false negative (classes that are wrongly classified as negative). 
\begin{equation}
\text { Accuracy }=\frac{T P+T N}{T P+T N+F P+F N}.
\label{m_acc}
\end{equation}
Equation~\ref{m_f1} denotes the calculation of F1-score, commonly used when there is class label imbalance, and both classes hold the same importance in classification metric. In Equation \ref{m_f1}, precision is the measure of the proportion of positive identifications that are actually correct, and recall is the measure of the proportion of actual positives that are correctly identified. 
\begin{equation}
    \text{F1-score} = 2 \times \frac{\text{precision} \times \text{recall}}{\text{precision} + \text{recall}}.
    \label{m_f1}
\end{equation}
To assess the DL model for regression problems, metrics like max variance, R-square, root mean squared error (RMSE), etc. are evaluated~\cite{kumar2021predictive,violos2021predicting,shitole2021optimization,moursi2021iot,farooq2021intelligent}.
RMSE defined in equation~\ref{m_rmse}, is a widely used metric for regression-based problems. In equation~\ref{m_rmse}, $y_i$ and $\hat{y_i}$ are the actual and predicted labels, and N is the total number of samples.
\begin{equation}
    \text{RMSE} = \sqrt{\frac{1}{N} \sum_{i =1}^N (y_i - \hat{y_i})^2}.
    \label{m_rmse}
\end{equation}
While these metrics are widely used, they are essential for the performance comparison of different models' architecture and strategies deployed on the same dataset over the ES.

\subsection{Training loss}
The process of training a DL model requires the optimization (typically minimization) of a specific loss function. The training loss is a metric that captures how well a DL model fits the training data by quantifying the loss between the predicted output and ground truth labels. Different metrics are selected based on the type of problem, \emph{i.e.}, classification or regression. Some of the widely used loss functions to capture the learning of the DL model at the edge while training are mean absolute error \cite{wang2019ecass,gao2020salient,figetakis2021uav}, mean square error \cite{zhu2021learning,yang2021EDGE}, negative log-likelihood \cite{shao2022task,liu2021resource}, cross-entropy \cite{li2021intelligent,du2021cracau,fagbohungbe2021efficient}, Kullback-Leibler divergence \cite{deng2021share,goldstein2021decentralized,sun2021cooperative} etc.\par 
Cross-entropy, also called logarithmic loss, log loss, or logistic loss, is a widely accepted loss function for classification problems. In cross-entropy, the predicted class probability is compared to the actual class label. A loss is calculated that penalizes the DL model higher if the probability is very far from the actual value. The penalty itself is logarithmic, which yields a significant score for large differences close to 1 and small score for small differences tending to 0.
The cross-entropy loss function is defined as
\begin{equation}
L(y, \hat{y})=-\sum_{i=1}^{n} y_{i} \log \left(p_{i}\right), \text { for } \mathrm{n} \text { classes, }
\end{equation}
where $p_{i}$ is the predicted class probability of the $i^{t h}$ class.
Similarly, for regression problems, mean squared error (MSE) is the most commonly used loss function. The loss is the mean of the squared differences between true and predicted values across the dataset. MSE is defined as:
\begin{equation}
L(y, \hat{y})=\frac{1}{N} \sum_{i=0}^{N}\left(y-\hat{y}_{i}\right)^{2}.
\end{equation}

\subsection{Convergence rate}

When training a DL model, we typically monitor its loss until it reaches some measure of convergence. We would expect the loss to decrease until any further updates to DL model parameters will not change the test dataset inference made by the DL model, known as the convergence of the DL model.  
% The objective of DL model while training in a centralized fashion is to converge to an optimum solution. The optimum solution also means the convergence of the DL model, which can be measured by tracking the loss of the DL model. If the loss of the DL model is not improving, that DL model is assumed to have reached an optimum solution. 
The convergence rate is normally computed when using a distributed and decentralized architecture to train a DL model at the edge. One of the primary goals of the distributed/ decentralized DL model training at the edge is to speed up the convergence of DL models getting trained at multiple locations. Thus, DL models at different ESs need to collectively converge to a consensus that any further updates in the model will not change the estimate of the model for a given classification or regression problem \cite{guo2020communication}. Convergence rate, as a metric, defines the number of iterations one algorithm will take to converge to an optimum solution \cite{liu2020client}.  Thus, in a decentralized/ distributed architecture at all in-edge, the convergence rate as a metric becomes crucial because the different combinations of the architecture selected along with synchronization schemes (synchronous, asynchronous, etc.) have different convergence rates \cite{wu2020collaborate,nedic2020distributed,jiang2020skcompress,so2021codedprivateml}.

\subsection{Latency}
When inferring from a model at the edge, both the computational latency and communication latency become critical key performance metrics. Computational latency provides an estimate of the time that the DL model will require to process a query input and infer on the same \cite{yang2021joint,zeng2021energy,liu2021light}. Whereas communication latency provides an estimate of the time from when a query is sent from the origin server until the result is returned \cite{li2021slicing,zhang2021making,zhu2021network,shlezinger2021collaborative}. For mission-critical cases \cite{chen20213u}, DL models with low computational and communication latency are more favored. This metric becomes critical because one of the reasons to move from cloud to all in-edge is to reduce the latency incurred during the DL inference phase. The measuring unit of latency can range from milliseconds to seconds based on the latency requirement from DL-based applications. 

\subsection{Communication cost}
When a DL model is deployed for inference on an ES, many requests by the end user(s) are raised to get inference from the DL model. The volume of data, \emph{e.g.,} kilobytes (KB) or megabytes (MB), transmitted from the end user(s) has the potential to create congestion at the ES. The communication cost metric evaluates the amount of data (message size of each query) flowing to the ES from the end user \cite{shi2020communication,lim2021decentralized}. It also takes into consideration the inference data, which is reverted to the end user. Active monitoring of the communication cost is important to prevent potential congestion points \cite{welagedara2021review,janakaraj2021towards,tonellotto2021neural}. In typical cases, measuring the unit of communication cost is kept in KB or MB based on how much data is required to make an inference.
\iffalse
As a model is trained using a decentralized/ distributed architecture, it is necessary to transmit intermediary output between different partitions of the model, which are hosted across multiple ES. The available network bandwidth represents a constraint on the transmissions of these messages. Thereby the quantity/size of messages being passed between the ES can cause a bottleneck in the network. To avoid network congestion, the message size and frequency need to be evaluated while training the model.
\fi

\subsection{Privacy preserving}
\label{pm:utilityme}
Privacy-preserving metrics provide a means to quantify the level of user privacy offered by a DL model using privacy-preserving technologies \cite{wagner2018technical}. We can assess the ability of a model to retain data privacy during the training and inference phases. In both phases, there are two types of data leakage: direct and indirect. Direct leakage at the training phase occurs when an external party gains access to non-encrypted training data sent to a centralized ES. In addition, direct leakage can also occur when in a decentralized/distributed
setting when an external party gains access to activations or gradients that are sent from one edge server to another during training. Similarly, direct leakage at the inference phase occurs when an external party gains access to non-encrypted client data sent to an ES hosting the DL model. Indirect leakage at the training phase occurs when an external party gains access to DL model parameters which can indirectly provide information regarding training data. Indirect leakage from the inference phase comprises results provided by the DL model, which can leak sensitive information regarding the data DL model, is trained upon, i.e., membership inference attack and model inversion attack. Well-established encryption algorithms manage direct leakage from non-encrypted data at the training and inference phase, like DES, 3DES, AES, RSA, and blowfish, which don't require evaluation \cite{vaibhavi2021survey}.  \par
During the training phase, we can use the mutual information score (MIS) to measure the level of direct leakage (activation or gradients being sent from one edge server to another) or indirect leakage (access to DL model parameters)~\cite{joshi2022performance}.\par
The mutual information score (represented as $I$, in equation~\ref{m_mi}) measures how much information a random variable $X$ ({\emph{e.g.}}, smashed data/ model parameters) can reveal about another random variable $Y$ ({\emph{e.g.}}, non-encrypted training data/ another set of model parameters). For $X$ and $Y$ with joint distribution of $p(x,y)$, it is defined as follows:
\begin{equation}
   I(X, Y ) = \sum_{x\in X, y\in Y} p(x,y) log\frac{p(x,y)}{p(x)p(y)}.
    \label{m_mi}
\end{equation}
This metric ranges from 0 to 1, where 0 implies the raw data are independent of the intermediary activation vector or model parameters differ from another set of model parameters. \par
During the inference phase, two attacks (the model inversion attack and the memberships inference attack) can lead to indirect leakage. A model inversion attack allows an adversary to recover the confidential dataset utilized for training a supervised neural network. In an image-based model to evaluate model inversion, the structural similarity index measure (SSIM) is used to evaluate the reconstruction accuracy~\cite{subbanna2021analysis}. The magnitude of the deformation field resulting from non-linear registration of the original and reconstructed images is used to evaluate the reconstruction accuracy. The structural similarity index measure between two images $x$ and $y$ of common size $N\times N$ is:
\begin{equation}
\operatorname{SSIM}(x, y)=\frac{\left(2 \mu_x \mu_y+c_1\right)\left(2 \sigma_{x y}+c_2\right)}{\left(\mu_x^2+\mu_y^2+c_1\right)\left(\sigma_x^2+\sigma_y^2+c_2\right)},
\end{equation}
where
\begin{enumerate}
\item $\mu_x$ the average of $x$,
\item $\mu_y$ the average of $y$,
\item $\sigma_x^2$ the variance of $x$,
\item $\sigma_y^2$ the variance of $y$,
\item $\sigma_{x y}$ the covariance of $x$ and $y$,
\item $c_1=\left(k_1 L\right)^2, c_2=\left(k_2 L\right)^2$ two variables to stabilize the division with weak denominator,
\item $L$ the dynamic range of the pixel-values (typically this is $2^{\# \text { bits per pixel }}-1$ ), and
\item $k_1=0.01$ and $k_2=0.03$ by default. 
\end{enumerate}

A membership inference attack, in contrast, does not recover the training data but allows an adversary to query a deployed DL model to infer whether or not a particular example was contained in the model's training dataset. An adversary in this approach trains another DL model to infer whether a specific example was present in the training dataset. Accuracy as a metric is utilized to evaluate the quality of the adversary's DL model. One of the current metrics proposed to measure membership inference is the ratio of the true-positive rate to false-positive rates. This metric provides more strict measures to make the DL model provide a guarantee that in an ideal scenario, none of the positive cases should be incorrectly identified. This metric becomes different from AUC- ROC curve as TPR is only reported for fixed low FPR (\emph{e.g.}, 0.001\% or 0.1\%) \cite{carlini2022membership}.

\subsection{Energy consumption}
There is a wide range of available DL models, and their individual energy requirements for computation can vary significantly. For some resource-constrained environments, it becomes infeasible to host models with a larger energy footprint \cite{desislavov2021analysis}. The energy requirements of different models should be evaluated for their training and inference phase at the all in-edge level\cite{nez2021dynamic,mei2021energy,zhu2021green}. Power consumption (watts and kilowatts units) as measurement can be utilized to determine energy consumption \cite{liang2020ai}. Clearly, this metric is particularly relevant to an ES when hosting all the parts of a deep learning model.

\subsection{Memory footprint/ Model size}
For an ES with limited computational resources, it can be challenging to host a DL model with a huge number of parameters. The larger the DL model the more parameters it will have, and consequently the more memory space (in RAM) required to host the model. Model size or memory footprint is computed having `MB' as their unit of measurement \cite{flamis2021best,varghese2021survey,chen2019neuropilot,merenda2020EDGE,liu2021bringing}. For a specific image classification problem, if MobileNet V2 with 3.54 million parameters is selected, it will have 14 MB as model size whereas if InceptionV4 with 42.74 million parameters is selected for the same problem, it will have a 163 MB model size requirement \cite{liang2020ai}.

\subsection{Combined metrics}
As all in-edge needs to satisfy multiple constraints (\emph{i.e.}, energy, quality of DL model, latency etc.), it becomes more important to introduce hybrid metrics that combine multiple metrics. For example, the energy-precision ratio (EPR)~\cite{epr}, provides a way to combine the classification error with the energy consumed per sample. In equation~\ref{m_epr}, energy-precision ratio (EPR) is defined as: 
\begin{equation}
    \text{EPR} = \text{Error}^\alpha \times \text{EPI},
    \label{m_epr}
\end{equation}
where Error is the classification error, $\alpha$ is the adjustment parameter, and EPI is the energy consumption per sample.

\subsection{Robustness}
Adversarial examples can manipulate DL models, and negatively affect the models' performance or lead to misclassification. Thus the models need to be either robust to these examples by default or integrate various defence techniques to strengthen their robustness properties. The robustness of a model is defined as the insensitivity of the model to small perturbations made to any plausible input.
Moreover, the robustness can be defined as the reciprocal of the KL Divergence:
\begin{equation}
    \psi (x) = \frac{1}{\max\limits_{\delta \in set} D_{\text{KL}} (\hat{y}, \hat{y'})},
\end{equation}
where $D_{\text{KL}}$ is the KL Divergence, $\hat{y}$ and $\hat{y'}$ are the predictions for a sample $x$ and $x+\delta$, respectively~\cite{robustnessdef}. Another simple way to measure the robustness can be the difference in the accuracy with and without the adversarial examples. 
The defence techniques for robustness include PixelDP, which is a certified defence for norm-bounded adversarial samples~\cite{certifiedrobust}, adversarial training and ensemble learning.  

\subsection{Transferability and lifelong learning}
The ability to reuse previously learned information for a related task indicates the transferability of a model. Thus, there is no need to train the model from scratch for a new task if the model is transferable from another related domain~\cite{deeptransfer}.  
The transferability can be measured by using transfer accuracy and LEEP score ($T(\theta, D)$)~\cite{leepscore}.  
\begin{equation}
    T(\theta, D) = {\frac{1}{n}} \sum_{i=1}^n( \sum_{z\in Z} \hat{P}(y_i|z), \theta(x_i)_z),
\end{equation}
where $x_i$ is a sample, $\theta$ is the source model, $D$ is the target dataset, $z \in Z$ is the label in the label set $Z$ of the source task, and $\hat{P}(y_i|z)$ is the conditional distribution of the predicted labels given an original label. 

The environment can change over time, and the model needs to adjust accordingly to capture the changes. Thus, models capable of lifelong learning are preferable. In lifelong learning, the model retains the previously gained knowledge and also keeps learning new information with time. Overall, transferability and lifelong learning capability make the DL models data and computation efficient.

%==========================================
\section{Open Challenges and future direction}
Thus far, we have discussed DL architectures, technologies, adaption techniques, and the key performance indicators required to facilitate DL to all in-edge. In this section, we now articulate the key open challenges and future research directions in the area of DL at all in-edge.
\subsection{Challenges with resource-constrained edge servers}
\label{openchallange1}
It's necessary to know the configuration of ESs before starting the training and deployment of the DL model at ESs. This section discusses challenges that arise from heterogeneous ESs provided by different edge infrastructure providers (\emph{e.g.}, Motorola Solutions, Hikvision, ADT) and associated future directions of research.

%\begin{enumerate}[leftmargin=*]
\paragraph*{1) Memory efficiency}

There are significant challenges to facilitating both the training and inference of DL models on ESs due to the limited resources and heterogeneous configuration of different ESs. DL models can vary significantly in their overall size. For example, inception-v3 has a size of 91 MB~\cite{bhatt2017comparison}, while vgg-19 has a size of 548 MB~\cite{wang2018bml}). Thus, based on the selected DL model (assuming it to be vgg-19)  and enabling technology (assuming it to be federated learning), it can become impossible for some ES to participate in DL model training due to insufficient memory (if memory is less than 548 MB). The lack of availability of certain ESs can negatively impact the DL model convergence rate (a small number of available ESs for distributed training will mean a slower convergence rate). Also, due to the fairly large size, some DL models will be limited to being deployed for inference at a small number of ESs. In the future, explore the direction of utilization of heterogeneous ESs by answering: How can we train a DL model at ESs where some models can train fairly large models due to extensive memory, and some can partially train those models? How can we design DL models to facilitate training across heterogeneous ESs? 

\paragraph*{2) Energy requirement}

As ESs in remote locations can be battery-powered, minimizing energy consumption is a critical ongoing challenge. One way to achieve it is by limiting the computation required in the training and inference phase, which inherently lowers the energy requirement. Another important avenue of research is to investigate the performance of battery-operated ESs when different DL models are trained and deployed. While chipset designers continuously strive to reduce the energy requirements of their products (GPUs, TPUs, etc.). The same understanding of the interaction of the rest of ES composition (computing chipset, storage drives, batteries, etc. are required) to find a fair trade-off between battery management and compute resources is required.
%\end{enumerate}

\subsection{Quality of Service (QoS) attributes for DL model at an all in-edge level}
\label{openchallange2}
In order to be competitive with a centralized cloud model, the all in-edge model needs to provide quality of service guarantees. This section discusses the "DL model at all in-edge guarantees" to build a complete all in-edge framework for DL.
%\begin{enumerate}
\paragraph*{1) Low Latency} 

Low latency is the first attribute that needs to be fulfilled at an all in-edge level. Low latency can be achieved by providing faster communication during model training and a quicker inference response from a deployed DL model. Due to the closer proximity of a deployed DL model to the end users, reduced latency has been observed in edge-based models compared to the traditional cloud-based models. For real-world applications, DL applications like image segmentation, object detection, etc., require very low latency. Using edge-based DL models, academic and industrial researchers actively seek ways to reduce latency \cite{li2019automatic,khani2021real,ilhan2021offloading,mahendran2021computer,rahman2021internet}. Although progress has been made in this area, the current state-of-art still results in significant latency, specifically when dealing with high dimensional input data (\emph{e.g.}, image, time series). For example, a constrained model architecture can process between 5 to 15 frames per second (fps) with an image resolution set to $1920 \times 1080$ \cite{george2021mez}. However, processing 5 to 15 frames per second is relatively lower than the fps at which videos are captured (typically 24 fps higher). Processing a higher number of frames will result in delayed latency at the inference stage; This remains an open research problem and as such new techniques are required to deal with high dimensional data.\par
Similarly, model compression provides approaches to reduce latency by enabling larger DL models to be deployed at an ES. This reduces the computation required (as the DL model is quantized and compressed). However, DL networks have continued to grow in size (leading to a corresponding increase in the number of parameters). This necessitates further research on providing more powerful compression techniques for DL networks.
\newpage

\paragraph*{2) Heterogeneous data distribution and asynchronous edge server participation}

The second attribute required by the DL model at the all in-edge level is its ability to be trainable at ESs with heterogeneous data distribution. Heterogeneous data distribution is caused by the non-Identical and Independent Distribution (non-IID) of the data among the multiple ESs, which leads to severe statistical heterogeneity challenges when training a DL model. For example, one extreme case is when an end server only has data from a particular class. Usually, DL algorithms trained in a distributed environment with multiple ESs with an overall non-IID data distribution will perform poorly~\cite {ma2022state}. This opens an interesting future direction of research. Adaptive optimization is one approach that can be used to improve the convergence speed of a DL model and can effectively mitigate the concerns of non-IID data distribution. For example, \cite{mills2019communication} proposed adapting FedAvg to use a distributed form of Adam optimization to implement adaptive federated learning, which converges to a target accuracy in $6 \times$ fewer rounds than compressed FedAvg. In the future, exploring momentum, adaptive optimization, learning rates, and other hyperparameters is a worthwhile research direction in the context of a non-IID distribution. In addition, the participation of the ESs can be inconsistent due to communication or computational reasons. This results in either a slowdown in the convergence rate or an inability to converge at all~\cite{kairouz2021advances},  thus placing more emphasis on asynchronous ES participation in the training phase. Future research into the robustness of training models in such scenarios is also warranted. 
\iffalse
\item DL model design Vs key performance indicators: 

When selecting any DL model to cater to a specific mission, a series of DL model candidates can be reviewed as capable of completing the task. Although academics and industrial papers evaluate and compare the DL models, they are often limited to some DL models considered for that research. However, a drawback of limited comparisons gets amplified when these researches are accompanied by a different set of performance indicators such as top-k accuracy~\cite{khoa2021fed} and mean average precision~\cite{yang2022distilled}. Unfortunately, these standard metrics fail to provide insights into the runtime performance of DL model inference at ESs. Besides mission-specific performance indicators(\emph{e.g.}, accuracy), other key performance indicators such as convergence rate, latency, communication cost, energy consumption, memory footprint, etc., are also crucial for the DL model deployed at the ES. However, this survey paper captures key performance indicators that should be provided (in Section ~\ref{sec:util}). We need to identify more combined metrics which can quantitatively analyze the trade-off between independent indicators to help improve the deployment of the DL model at the ES.
\fi
%\end{enumerate}

\subsection{Privacy and security concerns}
\label{openchallange3}
Despite the rapid development of privacy-preserving DL~\cite{li2020preserving} and security mitigation techniques~\cite{rahman2022deep} in recent years, there are still open research challenges that need to be addressed. This section discusses potential open research problems and future directions regarding privacy and security concerns impacting DL model development and deployment at all in-edge.
%\begin{enumerate}
\paragraph*{1) Privacy-preservation}

Providing adequate privacy preservation for DL applications is an area with open research challenges. To preserve the privacy of the client's data at the ES, different enabling technologies are utilized with or without cryptographic techniques, perturbation techniques, and anonymization techniques \cite{shokri2021privacy,tran2019privacy}. While these techniques provide a means of better safeguarding client data, they struggle to maintain the original level of model performance simultaneously. For example, the inclusion of these techniques can not only negatively impact the predictive performance of a model (accuracy, F1-score, etc) \cite{kaissis2021end,alkhelaiwi2021efficient,gawali2021comparison} but can also significantly lengthen the training \cite{zhang2021adaptive,thapa2021advancements} and inference \cite{jain2022ppdl,tan2021cryptgpu}) time of a model. Therefore, there are opportunities in this area to preserve privacy while mitigating the negative consequences outlined above. 

\paragraph*{2) Security} 

The ESs need active participation while enabling DL at the all in-edge level. However, due to hardware constraints (\emph{e.g.}, low computational capability) and software heterogeneities of the ESs, this also represents an increase in the attack surface.  Moreover, various attacks such as Distributed Denial-of-service(DDoS) targeting network/virtualization infrastructure, side-channel attacks targetting user data/privacy, malware injection targeting ES/devices, authentication and authorization attacks targeting ES/devices and virtualization infrastructure are possible for all in-edge computing system~\cite{xiao}. However, finding efficient and suitable countermeasures for these attacks is challenging due to constantly evolving attackers' tactics, techniques, and procedures~\cite{xiong2022cyber}. Besides, DL approaches such as federated learning and split learning for edge intelligence suffer from adversarial attacks on the federated models to modify their behavior and extract/reconstruct original data~\cite{splitattack}. To that end, novel techniques are required to identify security attacks/breaches and mitigate such attacks in the future.

%\end{enumerate}

\subsection{Framework and architectural changes to facilitate DL models at all in-edge level}
The convergence of DL at the all in-edge level is a relatively new paradigm, with concerns about effective resource utilization, management, and interoperability amongst heterogeneous ESs, requiring new frameworks and architectural changes. This section will discuss promising directions that can help mitigate the concerns at the convergence.
%\begin{enumerate}
\paragraph*{1) Microservices}

As computing is getting pushed away from being cloud-based to edge-based, architectures to facilitate DL model training and deployment are also shifting from monolithic entities to graphs of loosely-coupled microservices~\cite{al2022ai}. Microservices provide a promising way of modularizing DL-based applications at the process level. For example, a single DL application can be decomposed into a non-overlapping atomic set of services in a microservice architecture. However, sometimes one DL model inference can depend on another DL model inference. At the same time, another DL model may require different computing languages (\emph{e.g.}, python, R), language dependencies (\emph{e.g.}, PyTorch, TensorFlow), and software dependencies (\emph{e.g.}, pycharm, GitLab). Microservices architectures provide a means for those DL models with different requirements to communicate effectively. Currently, the introduction of microservices for deploying and training at the edge is at a very early stage~\cite{yang2021edgekeeper}. The research opportunity exists to build a robust microservice framework that can handle the deployment and management of the DL model. Another opportunity lies in migrating microservices-based DL applications from development to production with minimal downtime.  

\paragraph*{2) Management of DL-based applications at ESs}

The confluence of DL models deployed at ESs and the emergence of smart cities has led to a new interesting research area of DL-assisted smart cities. With many DL models deployed at ESs in smart cities, it will become challenging to predict the future requirement of resources for DL computation accurately. Real-time optimization will be required amongst ESs to accommodate heterogeneous computation and communication adaptively. As a result, better resource orchestrators (online ES management applications)  will be required at the edge to facilitate the potentially large number of requests that will be generated within an ecosystem of smart cities. Also, with every government taking steps toward smart cities, these orchestrations will be dispersed across different geolocations and regions, thus providing an opportunity for collaboration between individual orchestrators. A flexible coordination mechanism between orchestrators situated adjacent to each other will be required, which can also preserve citizens' privacy. An emerging research direction is utilizing AI to tackle the design complexity of interconnected smart cities; One of the ways to achieve it is by using deep reinforcement learning (DRL) \cite{gong2021distributed}. A distributed DRL-based scheme can provide an efficient way to solve the data-driven interference mitigation and resource allocation problem. It also opens up new research opportunities on the need to develop a uniform API interface for ubiquitous heterogeneous ESs to ease the deployment of orchestrators. Due to the highly dynamic nature of this environment (any ES can go offline and come back into service), an important and related research direction is the design of efficient service discovery protocols. Service discovery protocols will provide necessary information to companion ESs regarding what can be expected from DL-based applications deployed at that ES.

\paragraph*{3) Designing application framework to facilitate DL at the all in-edge level}

All in-edge paradigm requires new ways of designing applications. In Section \ref{modelTrainingArch}, we presented different architectures capable of pushing AI to the ES with varying application requirements. With the enabling technologies explained in Section \ref{EnablingTechnologies}) and model adaption techniques described in Section \ref{modelAdaption}, developing DL applications becomes progressively more complex. The aforementioned microservices-based architecture is another exciting area of research in the provisioning of DL-based applications at ESs \cite{ezzeddine2018restful}. Although other research provided the framework for designing DL-based applications by utilizing ESs, they all remain confined to the problem they tried to resolve. For example, in \cite{xiao2020toward} provided a framework for the self-learning DL model, in which authors proposed a GAN-based synthesis of the traffic images. The proposed framework remains applicable only for video-based scenarios.
Similarly, the work in \cite{an2020novel} provides a framework that was restricted to work for web traffic anomaly detection. Likewise, other research \cite{li2018EDGE,liu2019e2m} has its niche, and the proposed framework is restricted to solving the specific problem type. To the author's knowledge, Open EI \cite{zhang2019openei} is the only framework that provides a generic approach to facilitate the development of applications for a wide range of problem domains (computer vision, natural language processing, etc.). Still, this framework lacks the components of hardware (choices in the selection of hardware accelerators that can help in faster DNN computation \cite{hashemi2021darknight,shehzad2021scalable,zaman2021custom,mittal2021survey}) and the deployment of the DL-based services (how to distribute load and develop a global model across the ES \ref{EnablingTechnologies}). Therefore, there is a need to find a robust framework that can facilitate the easy development and deployment of complex DL-based applications at the all in-edge level by providing guarantees from DL-based applications (as mentioned in Section~\ref{openchallange2}) while adhering to infrastructural constraints of the ES resources (as discussed in Section~\ref{openchallange1}) alongside mitigating the privacy and security concerns (as described in Section~\ref{openchallange3}).

% \end{enumerate}

\section{Conclusion}
This paper reviewed the current states to facilitate the training and inference of DL models on a fine mesh of ESs (referred to as all in-edge level). The behavior of centralized, decentralized, and distributed architecture were discussed from the ES's perspective to find a trade-off between simplicity (by centralized architecture) or achieving reliability (by utilizing a decentralized and distributed architecture) for DL models deployed at the all in-edge level. Technologies facilitating the DL training and deployment across ESs were described, which leverage the layer structure of the DL models and closer proximity to the origin of the data. Federated learning and split learning as enabling technologies were more effective than others as they provided enhanced privacy while training and providing inference from the DL model. Model adaption techniques were found to be necessary at all in-edge, providing benefits of minimizing the energy requirement, lowering communication message size, and decreasing the memory footprint. In addition to general performance indicators, this paper identified and put forward additional key performance indicators, measured in silos in several works but not considered to be evaluated simultaneously. Many research directions remain open regarding optimizing memory and energy of resource-constrained ESs for facilitating DL at ESs while preserving the privacy of the user's data, incorporating advancements in cybersecurity to diminish security concerns, and lastly, close collaboration with networking technologies (such as network functions virtualization). With new technological innovations, shifts in DL-based application design, networking technologies improvements, and ESs hardware advances, many of the previously mentioned challenges will be mitigated. This will bring new challenges and opportunities for further innovation.

\section*{Acknowledgment}

This research was conducted with the financial support of ADVANCE CRT
PHD Cohort under Grant Agreement No. 18/CRT/6222 and at the ADAPT
SFI Research Centre at Cork Institute Of Technology. The ADAPT SFI Centre
for Digital Media Technology is funded by Science Foundation Ireland through
the SFI Research Centres Programme and is co-funded under the European
Regional Development Fund (ERDF) through Grant 13/RC/2106.

%% ----------------------------------------------------------------

\bibliographystyle{ACM-Reference-Format}  % Use the "IEEE Transaction" BibTeX style for formatting the Bibliography

  % The references (bibliography) information are stored in the file named "Bibliography.bib"
%% ----------------------------------------------------------------

\end{document}

%% file: preamble.tex
\usepackage{xspace}
\usepackage{pifont}
\usepackage{tikz}
\usepackage{pgfplots}
\usepackage{multirow}
\pgfplotsset{compat=newest}
\usepackage{lscape}